%% file: main.tex
\documentclass[conference]{IEEEtran}
\input{packages}
\input{commands}

\IEEEoverridecommandlockouts
\def\BibTeX{{\rm B\kern-.05em{\sc i\kern-.025em b}\kern-.08em
    T\kern-.1667em\lower.7ex\hbox{E}\kern-.125emX}}
\begin{document}

\title{Impact of Facial Tattoos and Paintings on Face Recognition Systems\\
{\large Mathias Ibsen, Christian Rathgeb, Thomas Fink, Pawel Drozdowski, Christoph Busch }
\thanks{The authors are with the da/sec – Biometrics and Internet Security Research Group, Hochschule Darmstadt, Germany.\\
E-mail: {mathias.ibsen}@h-da.de}
}
\author{}
\maketitle
\thispagestyle{plain}
\pagestyle{plain}

\begin{abstract}
In the past years, face recognition technologies have shown impressive recognition performance, mainly due to recent developments in deep convolutional neural networks. Notwithstanding those improvements, several challenges which affect the performance of face recognition systems remain. In this work, we investigate the impact that facial tattoos and paintings have on current face recognition systems. To this end, we first collected an appropriate database containing image-pairs of individuals with and without facial tattoos or paintings. The assembled database was used to evaluate how facial tattoos and paintings affect the detection, quality estimation, as well as the feature extraction and comparison modules of a face recognition system. The impact on these modules was evaluated using state-of-the-art open-source and commercial systems. The obtained results show that facial tattoos and paintings affect all the tested modules, especially for images where a large area of the face is covered with tattoos or paintings. Our work is an initial case-study and indicates a need to design algorithms which are robust to the visual changes caused by facial tattoos and paintings.\newline
\end{abstract}
\input{sections/introduction}

\input{sections/related_work}

\input{sections/database}

\input{sections/experiments}

\input{sections/results}

\input{sections/discussion}

\input{sections/conclusion}

\input{sections/acknowledgements}

\bibliographystyle{IEEEtran.bst}
\bibliography{bibli}
\end{document}

%% file: packages.tex
\usepackage{subcaption}
\usepackage{url}
{}
{}
{}
\usepackage{float}
\usepackage{multirow, multicol}
\usepackage{graphicx}
\usepackage{hhline}
\usepackage{amsmath}
\usepackage{array, makecell}%
\usepackage{bm}
\usepackage{amsbsy}
\usepackage{booktabs}
\usepackage{url}

\usepackage{hyperref}
\usepackage{breakurl}
\hypersetup{breaklinks,hidelinks}
\usepackage{xspace}
\usepackage{xcolor}
\usepackage{newfloat}
\usepackage[export]{adjustbox}
\usepackage{enumitem}

%% file: commands.tex
\DeclareMathAlphabet{\pazocal}{OMS}{zplm}{m}{n}
\newcommand{\de}{$\mathcal{D}_\xi$}
\newcommand{\n}{\pmb{\pazocal{N}}}
\newcommand{\tp}{\pmb{\pazocal{T}\pazocal{P}}}
\newcommand{\tps}{\pmb{\pazocal{T}\pazocal{P}_{small}}}
\newcommand{\tpm}{\pmb{\pazocal{T}\pazocal{P}_{medium}}}
\newcommand{\tpl}{\pmb{\pazocal{T}\pazocal{P}_{large}}}

\newcommand{\g}{\pmb{\pazocal{G}}}
\newcommand{\ig}{\pmb{\pazocal{I}}}

\newcommand{\eer}{$\mathit{EER}$}
\newcommand{\fnmr}{$\mathit{FNMR}$}
\newcommand{\fmr}{$\mathit{FMR}$}
\newcommand{\fte}{$\mathit{FTE}$}
\newcommand{\frr}{$\mathit{FRR}$}

\newcommand{\cots}{\textit{COTS}\xspace}
\newcommand{\serfiq}{\textit{SER-FIQ}\xspace}
\newcommand{\dlib}{\textit{dlib}\xspace}
\newcommand{\faceqnet}{\textit{FaceQnet}\xspace}
\newcommand{\arcface}{\textit{ArcFace}\xspace}
\newcommand{\mtcnn}{\textit{MTCNN}\xspace}

%% file: sections/introduction.tex
\section{Introduction}
Face recognition is used in a range of security applications for automatically recognising individuals and has been an active research area for decades \cite{Zhao-FaceRecognSurvey-2003} \cite{Abate-2DAnd3DFaceRecognition-2007} \cite{LiJain-HandbookOfFaceRecognition-2011}. Towards implementing robust and reliable face recognition systems several factors have been identified which can negatively affect the performance of face recognition systems, \textit{e.g.}\ changes in age, facial pose, and facial expression \cite{LiJain-HandbookOfFaceRecognition-2011} \cite{Wang-DecorrelatedAdversarialLearning-IEEE-CVPR-2019}. Additionally, it has been shown that facial changes caused by different facial manipulations like plastic surgery, makeup, or facial retouching, might have a negative effect on the performance of face recognition systems \cite{Rathgeb-ImpactDetectionFacialBeautificationSurvey-ACCESS-2019}. It is therefore of interest whether other types of face manipulations, in particular, \textit{facial tattoos} and \textit{facial paintings}, have an impact on the biometric performance of face recognition systems. Buttle and East \cite{Buttle-TraditionalFacialTattoosDisruptFaceRecognitionProcesses-Perception-2010} conducted a study on the impact of facial tattoos on human perception; they showed a strong deterioration of the humans' ability to recognise faces with traditional spiral-like tattoos. However, to the best of our knowledge, there exists no work which investigates the impact of facial tattoos or paintings on automated face recognition systems.

\par During the enrolment phase of a classical face recognition system, a reference image is captured by the capture device whereafter the face is detected, alignment is performed, and sample quality is estimated. Lastly, a biometric feature set useful for face recognition is extracted and stored as a reference template in an enrolment database. At the time of authentication, a probe face image is captured and processed in the same way as during enrolment; subsequently, it is compared against a reference template corresponding to a claimed identity (verification) or up to all stored reference templates (identification). Traditionally, handcrafted texture descriptors such as Local Binary Patterns (LBP) \cite{Ahonen-FaceRecognitionWithLocalBinaryPatterns-ECCV-2004} \cite{Ahonen-FaceDescriptionWithLocalBinaryPatterns-IEEE-2006}, Histogram of Oriented Gradients (HOG) \cite{Deniz-FaceRecognitionUsingHistogramsOfOrientedGradients-2011}, and Gabor filters \cite{Zhang-LocalGaborBinaryPatternHistogramSequence-IEEE-2005}, have been used for extracting features useful for facial recognition. The reader is referred to \cite{Zhao-FaceRecognSurvey-2003} \cite{Abate-2DAnd3DFaceRecognition-2007} \cite{LiJain-HandbookOfFaceRecognition-2011} for an overview of such systems. Current state-of-the-art face recognition systems, however, take advantage of the availability of large 2D face image databases and deep learning techniques to learn complex representations of faces. This paradigm shift has led to an impressive improvement of biometric performance results of facial recognition systems \cite{Wang-DeepFaceRecognitionASurvey-Arxiv-2018} \cite{Ranjan-DeepLearningSurvey-IEEE-2018}  \cite{Deng-ArcFace-IEEE-CVPR-2019} \cite{Zhou-NaiveDeepFaceRecognitionTouchingTheLimit-2015-arxiv}. 

\begin{figure}[ht]
    \centering
    \includegraphics[width=1.05\columnwidth]{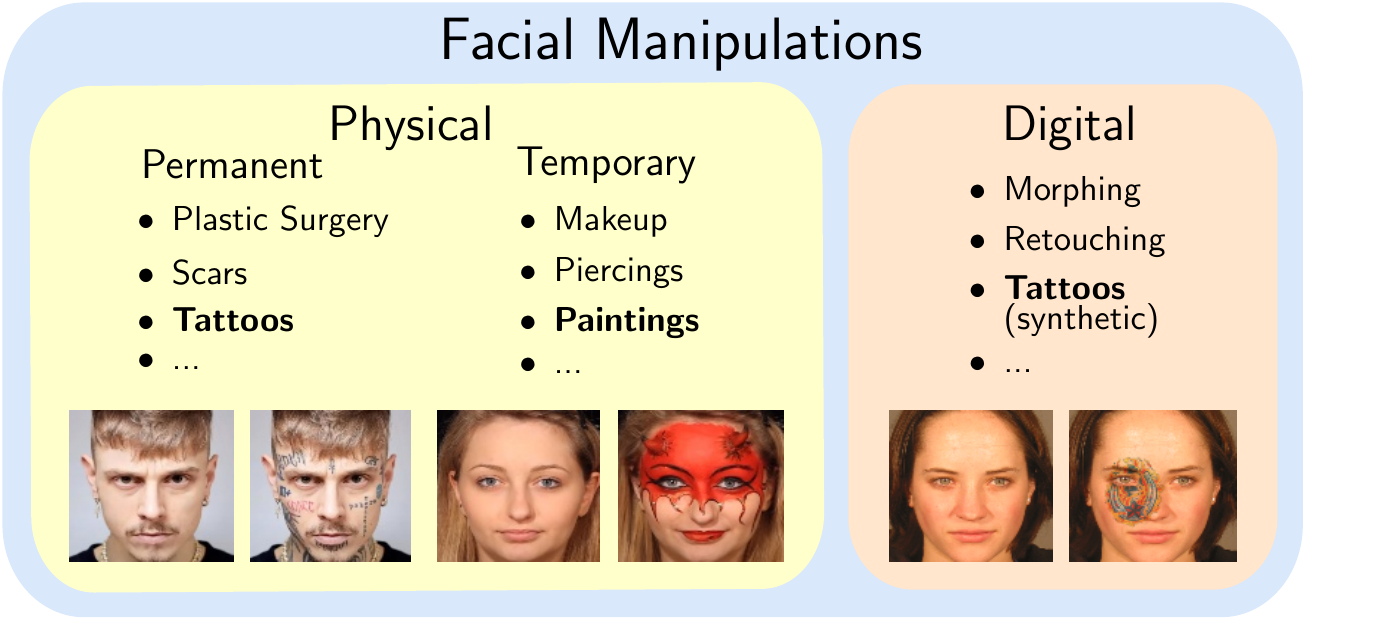}
    \caption{Overview of the main types of facial manipulations. Example images are shown for the facial manipulations highlighted in bold.}
    \label{fig:face_alterations_overview}
\end{figure}

Figure \ref{fig:face_alterations_overview} gives an overview of different types of facial manipulations. Recent research has shown that the efficacy of facial recognition systems is affected by certain types of manipulations (\textit{e.g.} morphing \cite{Scherhag-MorphingAttacks-Survey-IEEEAccess-2019}, retouching \cite{Rathgeb-ImpactDetectionFacialBeautificationSurvey-ACCESS-2019}, and makeup \cite{Rathgeb-MakePresentationAttacksReviewAndBenchmark-2020-IEEE}). Motivated by the need for deploying robust and reliable systems, we investigate the impact of tattoos and paintings on face recognition systems.
\raggedbottom

\begin{itemize}
    \item A \textit{facial tattoo} is a form of permanent facial manipulation induced by inserting ink into the dermis layer of skin on a person's face. The main intent of tattoos concerns aesthetic or descriptive purposes, but they have historically also been used for various functional purposes such as identification \cite{auschwitz_tattoo_identification}. Other notable uses of tattoos include permanent makeup, which is a cosmetic technique where tattoos are employed on the skin to resemble makeup; however, permanent makeup is not considered in this work.
    \item A \textit{facial painting} is a non-permanent manipulation where paint is applied to a face. As with tattoos, facial paintings are often intended for descriptive or aesthetic purposes; they are commonly seen at festivals and sports events.
\end{itemize}
\raggedbottom
In this work, focus is put on the scenario in which a face recognition system is presented with an image pair to be compared where \emph{either the reference or probe image} has been altered by facial tattoos or paintings.  
It has been noted that tattoos and paintings are expected to have an impact on face recognition systems \cite{Rathgeb-ImpactDetectionFacialBeautificationSurvey-ACCESS-2019}, but this topic has seen limited attention in the biometric scientific community. Suppose face recognition is applied to images where one or more individuals have facial tattoos or paintings. In such cases, the facial manipulations caused by tattoos or paintings likely affect face recognition. As such, it is of interest to investigate the impact these manipulations have on different modules of face recognition systems which in turn affects different scenarios where these systems are used.

\begin{description}
\item[\textbf{Face detection}] For facial tattoos and paintings which cause severe facial changes, it is likely that they affect the ability of a face recognition system to detect the region of interest, \textit{i.e.} the faces. If so, it is possible that tattoos and paintings can be used as a presentation attack for identity concealment for facial images captured in unsupervised acquisition conditions. Furthermore, if these manipulations affect face detection, this impact would be propagated to subsequent modules of the face recognition pipeline. 
\end{description}

\begin{description}
\item[\textbf{Face quality estimation}] Another relevant scenario to consider is the effect of facial tattoos and paintings on face quality estimation. Assume, for instance, that facial tattoos and paintings have a low quality score compared to similar face images without any facial manipulations. Consequently, in systems where automated quality estimation is used prior to enrolment into the database, it might not be possible for all individuals with facial tattoos or paintings to be enrolled into the system. Considering how widespread facial recognition is in both governmental (\textit{e.g.} border control) and personal security applications (\textit{e.g.} smartphone applications), ensuring inclusiveness and accessibility for everyone irrespective of individual appearance is an important concern. 
\end{description}

\begin{description}
\item[\textbf{Feature extraction and comparison}] Lastly, it is of interest to test whether the visual changes caused by facial tattoos and paintings have an impact on biometric recognition performance. This is relevant for all practical applications of facial biometrics; for example, in a criminal investigation when a face of a suspect with facial tattoos is compared to an old stored photo of the same suspect without tattoos. As such, it is interesting to assess whether facial recognition systems are capable of reliably extracting features and distinguishing individuals with facial tattoos and paintings.
\end{description}

To address the use-cases mentioned above, a database consisting of 500 image-pairs of individuals with and without facial tattoos or paintings was assembled from online sources. Thereafter, state-of-the-art commercial and open-source systems were used in an empiric evaluation in accordance with the current practices and international standards. The obtained results indicate that especially large facial tattoos and paintings have a significant impact on the performance of face recognition systems. To facilitate reproducible research, the used database is made available \cite{hdaTattooPaintDatabase}. The work conducted in this paper is, to our knowledge, the first of its kind.

The remainder of this paper is organised as follows: Section \ref{sec:related_work} provides relevant background information and related works. Section \ref{sec:databases} presents the assembled database. Section \ref{sec:experiments} and \ref{sec:results} describe the conducted experiments and achieved results, respectively. Section \ref{sec:discussion} and \ref{sec:conclusion} conclude the work with a discussion of the obtained results and contributions.

%% file: sections/related_work.tex
\section{Related Work}
\label{sec:related_work}
The following subsections briefly summarise the most relevant related works with respect to facial manipulations in the physical (section \ref{sec:alt_real}) and digital domain (section \ref{sec:alt_digi}). 
\subsection{Physical Facial Manipulations}\label{sec:alt_real}
Most research concerning the impact of permanent facial manipulations in the physical domain has been focused on plastic surgery. Pioneering work in this field was done by Singh \textit{et al.} who created databases containing facial images of individuals before and after plastic surgery \cite{Singh-EffectOfPlasticSurgeryOnFaceRecognitionAPreliminaryStudy-IEEE-2009} \cite{ Singh-PlasticSurgeryANewDimensionToFaceRecogniton-ieee}. In \cite{Singh-PlasticSurgeryANewDimensionToFaceRecogniton-ieee}, the authors evaluated the impact of plastic surgery on six face recognition algorithms. They concluded that more work should be done towards designing algorithms capable of dealing with the appearance changes occurring as a result of a facial plastic surgery operation. More recently, Rathgeb \textit{et al.} \cite{Rathgeb-PlasticSurgeryDeepFace-CVPRW-2020} introduced a new database of plastic surgery images which are mostly compliant with the ICAO quality requirements \cite{ICAO-9303-p9-2015}. Using the new database as well as the database proposed in \cite{Singh-PlasticSurgeryANewDimensionToFaceRecogniton-ieee}, Rathgeb \textit{et al.}\ showed that the impact of plastic-surgery on state-of-the-art commercial and open-source face recognition systems is less significant than initially reported by other researchers. For a more detailed overview regarding the impact of plastic surgery on face recognition systems and the work conducted in this field, the reader is referred to \cite{Rathgeb-ImpactDetectionFacialBeautificationSurvey-ACCESS-2019}. 

More works addressed the impact of temporary face manipulations on face recognition. In \cite{Ueda-InfluenceOfMakeupOnFaceRecognition-Perception-2010}, Ueda and Koyama investigated the impact that makeup has on humans' ability to recognise faces and found that it significantly decreased, especially in the presence of heavy makeup. In \cite{Dantcheva-CanFacialCosmeticsAffectTheMatchingAccuracyOfFaceRecognitionSystems}, Dantcheva \textit{et al.} found that the presence of makeup negatively affects the accuracy of face recognition systems and concluded that more work was needed to address the challenges imposed by facial makeup. Similar findings were reported by Wang \textit{et al.}\ in \cite{Wang-Recognizing-HumanFacesUnderDisguiseAndMakeup-IEEE-2016}, where the authors investigated the impact of recognising human faces under disguise and makeup. Makeup is usually used for aesthetic purposes, but due to its ability to significantly alter the appearance of a face, attacks based on makeup can be effective against face recognition systems. In \cite{Chen-SpoofingFacesUsingMakeupAnInvestigativeStudy-IEEE-2017}, the authors showed that face recognition systems are vulnerable to makeup-induced presentation attacks with the aim of impersonation. In \cite{Rathgeb-MakeupAttackDetection-IWBF-2020} and \cite{Rathgeb-DetectionOfMakeupPresentationAttacksBasedOnDeepFaceRepresentations-Arxiv-2020} Rathgeb \textit{et al.}\ showed that such makeup presentation attacks of good quality could compromise the security of face recognition systems. Comparable work was done with the \textit{Computer Vision Dazzle Camouflage} concept initiated by Adam Harvey \cite{cvDazzle}, where hairstyles and makeup were used to prevent face detection, \textit{i.e.}\ presentation attacks with the aim of concealment. The idea of the concept is to use prior knowledge of a specific algorithm to design looks which prevent detection by that algorithm.

No dedicated research has investigated the impact of facial tattoos or paintings on face recognition systems. However, in \cite{Singh-RecognizingDisguisedFacesInTheWild-TBIOM-2019}, the authors investigated the impact of disguises on face recognition systems using a database which contains, among other types of disguises, facial paintings. The results show that even the top-performing algorithms that were tested perform poorly for strongly disguised faces. Notably, it was shown that a majority of genuine samples which were misclassified had occlusions over the face region and close to the periocular region. These findings are in agreement with other studies which have shown that partial facial occlusions negatively affect the performance of face recognition systems. In this context, it is believed that especially partially occluded faces captured in unconstrained environments will be a major challenge in the future \cite{Zeng-ASurveyOfFaceRecognitionTechniquesUnderOcclusion-Arxiv-2020}.

\subsection{Digital Facial Manipulations}\label{sec:alt_digi}
Initial research concerning the impact that digital face manipulations have on face recognition was carried out by Ferrara \textit{et al.}\ \cite{Ferrara-TheMagicPassport-IJCB-2014} \cite{Ferrara-FaceImageAlterations-Springer-2016} who, among other techniques, investigated the impact of beautification (retouching) and morphing. For facial retouching, the authors reported a significant drop in performance on all tested systems. These findings were confirmed by Bharati \textit{et al.} \cite{Bharati-DetectingFacialRetouchingUsingSupervisedDeepLearning-IEEE-2016} \cite{Bharati-DepographyBasedFacialRetouchingDetectionUsingSubclassSupervisedSparseAutoencoder}. In \cite{Rathgeb-PRNU-Retouching-Detection-BMT-2020}, Rathgeb \textit{et al.} showed that face recognition systems might be robust to facial images which have been digitally altered by the application of moderate facial retouching. The process of digitally editing attributes of a face has been eased by the creation of GAN-based algorithms which can automatically alter attributes of a face, \textit{e.g.} the age or hair of an individual. The STGAN approach presented in \cite{Liu-AUnifiedSelectiveTransferNetworkForArbitraryImageAttributeEditing-IEEE-2019} is an example of such an algorithm. For morphing, Ferrara \textit{et al.} showed that the tested face recognition systems were vulnerable to attacks based on morphed images \cite{Ferrara-TheMagicPassport-IJCB-2014}. Since the initial experiments, much research has aimed at detecting morphed images; the reader is referred to \cite{Scherhag-MorphingAttacks-Survey-IEEEAccess-2019} for a comprehensive survey. Besides the aforementioned, there exist several other types of digital facial manipulations, \textit{e.g.} entire face synthesis, identity swap, as well as attribute manipulation and expression swapping; interested readers are referred to  \cite{Tolosana-DeepfakesAndBeyondASurveyOfFaceManipulationAndFakeDetection-Fusion-2020} and \cite{Garrido-AutomaticFaceReenactment-IEEE-2014} for more details. The so-called ``deepfakes'', where neural networks are used to create fake images, can also be used to create manipulated biometric images \cite{Tolosana-DeepfakesAndBeyondASurveyOfFaceManipulationAndFakeDetection-Fusion-2020} \cite{Verdoliva-MediaForensicsAndDeepfakesOverview-2020-IEEE}. A study \cite{Korshunov-DeepFakesANewThreatToFaceRecognitionAssessmentAndDetection-Arxiv-2018} showed that deepfake videos pose a challenge for both face recognition systems and detection algorithms. Many researchers have undertaken the task of detecting manipulated biometric face images in recent years, and benchmarks have already been carried out, \textit{e.g.} in  \cite{Rossler-FaceForensicsLearningToDetectManipulatedFacialImages-ICCV-2019}.

%% file: sections/database.tex
\section{Facial Paintings and Tattoo Database}
\label{sec:databases}
\begin{figure}[!htb]
\centering
\begin{subfigure}[t]{.48\columnwidth}
\centering
\includegraphics[width=.48\columnwidth]{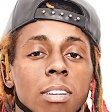}
\includegraphics[width=.48\columnwidth]{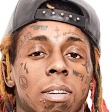}
\end{subfigure}\quad
\begin{subfigure}[t]{.48\columnwidth}
\centering
\includegraphics[width=.48\columnwidth]{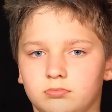}
\includegraphics[width=.48\columnwidth]{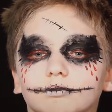}
\end{subfigure}
\begin{subfigure}[t]{.48\columnwidth}
\centering
\vspace{0.1pt}
\includegraphics[width=.48\columnwidth]{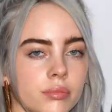}
\includegraphics[width=.48\columnwidth]{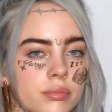}
\caption{Image pairs where one image in a pair has facial tattoos}
\end{subfigure}\quad
\begin{subfigure}[t]{.48\columnwidth}
\centering
\vspace{0.1pt}
\includegraphics[width=.48\columnwidth]{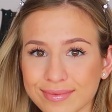}
\includegraphics[width=.48\columnwidth]{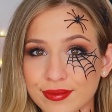}
\caption{Image pairs where one image in a pair has facial paintings}
\end{subfigure}
\caption{Examples of image pairs in the collected database.}
\label{fig:image_pairs_examples}
\end{figure}

While numerous 2D face databases have been collected for the purpose of evaluating face recognition systems \cite{Castaneda-ASurveyOf2DFaceDatabases-IEEE-2015}, there exists no publicly available database which can be used to investigate the impact of facial tattoos and paintings on face recognition systems. Therefore, an appropriate database was assembled using face images found online with Google image search and on YouTube. The collected database contains 500 pairs of face images of individuals with and without facial tattoos or paintings. Each image contains only a single visible face and by visual inspection, it has been ensured that there is not a significant age gap between two face images in the same pairing. Examples of image pairs in the assembled database are given in figure \ref{fig:image_pairs_examples}.

\begin{figure}[!htb]
\centering
\begin{subfigure}[t]{.48\columnwidth}
\centering
\includegraphics[width=.48\columnwidth]{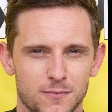}
\includegraphics[width=.48\columnwidth]{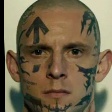}
    \caption{Semi-permanent facial tattoo with a real before image}
\end{subfigure}\quad 
\begin{subfigure}[t]{.48\columnwidth}
\centering
\includegraphics[width=.48\columnwidth]{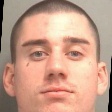}
\includegraphics[width=.48\columnwidth]{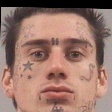}
\caption{Real tattoo with real before image.}
\end{subfigure}

\begin{subfigure}[t]{.48\columnwidth}
\centering
\includegraphics[width=.48\columnwidth]{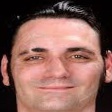}
\includegraphics[width=.48\columnwidth]{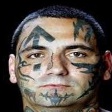}
    \caption{Real tattoo with real after picture (surgical removal)}
\end{subfigure}\quad
\begin{subfigure}[t]{.48\columnwidth}
\centering
\includegraphics[width=.48\columnwidth]{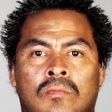}
\includegraphics[width=.48\columnwidth]{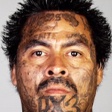}
\caption{Real tattoo with an image where tattoos have been digitally removed}
\end{subfigure}
\caption{Example image pairs illustrating the scenarios considered during the data collection.}
\label{fig:tattoo_scenarious}
\end{figure}

Four different scenarios were considered for the data collection of image pairs with facial tattoos, as illustrated in figure \ref{fig:tattoo_scenarious}.

\begin{figure}[htb]
    \centering %
\rotatebox[origin=c]{90}{\textsf{Tattoos}}\quad
\begin{subfigure}{0.2304\columnwidth}
  \includegraphics[width=\columnwidth]{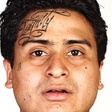}
  \label{fig:1}
\end{subfigure}\quad %
\begin{subfigure}{0.2304\columnwidth}
  \includegraphics[width=\columnwidth]{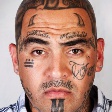}
  \label{fig:2}
\end{subfigure}\quad %
\begin{subfigure}{0.2304\columnwidth}
  \includegraphics[width=\columnwidth]{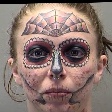}
  \label{fig:3}
\end{subfigure} \\
\rotatebox[origin=c]{90}{\textsf{Paintings}}\quad
\begin{subfigure}{0.2304\columnwidth}
  \includegraphics[width=\columnwidth]{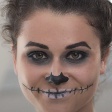}
  \caption{Small}
  \label{fig:4}
\end{subfigure}\quad %
\begin{subfigure}{0.2304\columnwidth}
  \includegraphics[width=\columnwidth]{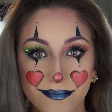}
  \caption{Medium}
  \label{fig:5}
\end{subfigure}\quad %
\begin{subfigure}{0.2304\columnwidth}
  \includegraphics[width=\columnwidth]{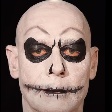}
  \caption{Large}
  \label{fig:6}
\end{subfigure}
 \caption{Example images showing small, medium, and large coverage of facial tattoos and paintings.}
\label{fig:different_coverage_ex}
\end{figure}

The collected database has been balanced such that there is a total of $250$ face images with facial tattoos and $250$ with face paintings; $232$ of the image pairs are of males, and the remaining $268$ images are of females. Out of the $250$ images with facial tattoos, $73.6\%$ of the images belong to men and $26.4\%$ to females. On the other hand $80.8\%$ of the images with facial paintings belong to women. More details about the assembled database are given in table \ref{tab:collected_database_metainfo} and examples for small, medium, or large coverage are shown in figure \ref{fig:different_coverage_ex}. 

\begin{table}[!htbp]
    \centering
    \caption{Overview of the assembled database}
    \begin{tabular}{@{}lll@{}} \toprule 
    \makecell[c]{\textbf{Category}} & \makecell[c]{\textbf{Subcategory}}  & \makecell[c]{\textbf{Distribution}} \\ \midrule 
    Type & \makecell[l]{Tattoo \\ Painting} & \makecell[l]{50.0\%\\ 50.0\%} \\ \midrule
    Size & \makecell[l]{Small: $<$ $1/3$ facial coverage \\  Medium: $1/3$ - $2/3$ facial coverage \\ Large: $>$ $2/3$ facial coverage} & \makecell[l]{31.2\%\\ 39.4\%\\ 29.4\%} \\ \midrule
    Gender & \makecell[l]{Male \\ Female} & \makecell[l]{46.4\%\\ 53.6\%} \\ \midrule
    Race & \makecell[l]{Caucasian \\ Hispanic \\ Black \\ Indian \\ Others} & \makecell[l]{41.0\% \\ 23.0\% \\ 14.6\% \\ 16.8\% \\ 4.6\%} \\ \midrule
    Age & \makecell[l]{Very young: $<$ 10 years old \\ Young: 10 - 30 years old \\ Middle-aged: 30 - 60 years old \\ Old: $>$ 60 years old} & \makecell[l]{13.8\% \\ 53.0\% \\ 31.0\% \\ 2.2\%} \\
    \bottomrule
    \end{tabular}
    \label{tab:collected_database_metainfo}
\end{table}

%% file: sections/experiments.tex
\section{Experimental Setup}
\label{sec:experiments}
The purpose of the experiments was to determine how facial tattoos and paintings influence algorithms ability to detect faces in an image, estimate face quality, as well as extract and compare face features. Table \ref{tab:modules_algortihms} lists which algorithms were used to test each of the aforementioned modules. For the evaluation, the database described in section \ref{sec:databases} was used. Additionally, for the comparison and feature extraction module, subsets of constrained images from the FRGCv2 \cite{Phillips-FRGC-2005} and FERET \cite{Phillips-FERET-1998} database were used.

\begin{table}[!htb]
    \centering
    \caption{An overview of the tested face recognition modules and used algorithms. \cots refers to the used commercial-off-the-shelf system.}
    \begin{tabular}{@{}ll@{}} \toprule \textbf{Module} & \textbf{Algorithms}   \\ \midrule 
    Face detection & \dlib, \mtcnn, \cots \\
    Face quality estimation & \serfiq, \faceqnet v1\\
    Feature extraction \& comparison & \arcface, \cots \\
    \bottomrule
    \end{tabular}
    \label{tab:modules_algortihms}
\end{table}

The following subsections describe the experimental setup designed for evaluating the impact of facial tattoos and paintings on the detection (section \ref{subsec:exp_face_detection}), quality estimation (section \ref{subsec:exp_face_quality}), as well as the feature extraction and comparison (section \ref{subsec:face_verification}) modules of a face recognition system. In the following, let $\n$ be the images in the assembled database without tattoos or paintings, let $\tp$ be the facial images with tattoos or paintings, and let $\tps$, $\tpm$, and $\tpl$ be the subset of facial images with small, medium, and large facial tattoos or paintings, respectively. The obtained detection confidence scores, quality estimation scores, and comparison scores were normalised to the range [0,1]. For systems where the system produced a narrow range of scores, the results were scaled using min-max normalisation.

\subsection{Face Detection}
\label{subsec:exp_face_detection}
To test the impact of facial tattoos and paintings on face detection, two open-source algorithms (\dlib \cite{King-MachineLearningToolkit-2009}, \mtcnn \cite{Zhang-JointFaceDetectionAndAllignmentUsingMultitaskCascadedConvolutionalNetworks-2016}) and one commercial-off-the-shelf (\cots) system were used. The detection performance of the algorithms was measured on the $\n,\ \tp,\ \tps,\ \tpm$, and $\tpl$ subsets of the assembled database. To this end, the confidence value (response score) produced by each algorithm for a detected region was recorded. The confidence value measures the likelihood that a detected region contains a face and as such a region with a high confidence value is likely to contain a face; whereas a region with a low confidence score is not. In case multiple regions were detected by an algorithm for a given image, the region with the highest confidence value was used. The confidence values produced for each of the subsets were compared using descriptive statistics calculated for each considered subset. Moreover, the detection error (\de), which is the percentage of facial images in a considered subset for which the algorithm detected zero faces, was recorded for each subset.

\subsection{Face Quality Estimation}
\label{subsec:exp_face_quality}
Face quality estimation is used in a variety of application-areas to estimate the utility of a facial image for face recognition; this is especially useful in unconstrained scenarios where low-quality facial images unsuitable for face recognition are likely to be captured. The \faceqnet v1 \cite{HernandezOrtega-FQA-FaceQnetV1-2020} and \serfiq systems \cite{Terhorst-FQA-SERFIQ-CVPR-2020} were used for measuring the quality score of the facial images. For \faceqnet, the images in the assembled database were cropped using the boundary box of the face detected by \mtcnn. For \serfiq, the \mtcnn face detector was used to normalise the facial images. For both systems, the quality score was estimated for each face in the database where \mtcnn detected at least one face. In case multiple faces were detected, the face with the highest confidence score was used. Results were computed and compared for the $\n,\ \tp,\ \tps,\ \tpm$, and $\tpl$ subsets using descriptive statistics. As the \mtcnn detector was used for pre-processing for both \serfiq and \faceqnet, the errors were not reported as they were similar to the \de\ (see section \ref{subsec:exp_face_detection}) scores of \mtcnn.

\subsection{Feature Extraction and Comparison}
\label{subsec:face_verification}
The impact that facial paintings and tattoos have on a face recognition system's ability to extract and compare features was evaluated using state-of-the-art open-source (\arcface \cite{Deng-ArcFace-IEEE-CVPR-2019}) and commercial (\cots) systems. Both considered systems are based on a deep-convolutional network and \arcface has been trained on the MS1MV2 dataset \cite{Deng-ArcFace-IEEE-CVPR-2019}.

The following scenarios were considered:

\begin{itemize}
    \item $\pmb{\pazocal{G}_{\pazocal{T}\pazocal{P}}}$: Genuine comparisons where one image in a pairing is from $\n$ and the other from $\tp$ (see section \ref{sec:databases}).
    \item $\pmb{\pazocal{G}_{\pazocal{T}\pazocal{P}_{small}}}$: Genuine comparisons where one image in a pairing is from $\n$ and the other from $\tps$ (see section \ref{sec:databases}).
    \item $\pmb{\pazocal{G}_{\pazocal{T}\pazocal{P}_{medium}}}$: Genuine comparisons where one image in a pairing is from $\n$ and the other from $\tpm$ (see section \ref{sec:databases}).
    \item $\pmb{\pazocal{G}_{\pazocal{T}\pazocal{P}_{large}}}$: Genuine comparisons where one image in a pairing is from $\n$ and the other from $\tpl$ (see section \ref{sec:databases}).
    \item $\pmb{\pazocal{G}}$: Genuine comparisons where both images have no facial tattoos or paintings. Subsets of constrained facial images from the FRGCv2 and FERET image databases were used.
    \item $\pmb{\pazocal{I}}$: impostor comparisons where both images have no facial tattoos or paintings. Subsets of constrained facial images from the FRGCv2 and FERET image databases were used.
\end{itemize}

For each of the above scenarios, the similarity scores were computed for both \arcface and \cots. Table \ref{tab:comparisons_overview} shows the number of comparisons for each considered scenario. The obtained comparison scores were compared using descriptive statistics. Additionally, the equal error rate (\eer), failure-to-enrol rate (\fte), and false non-match rate (\fnmr) at a fixed false match rate (\fmr) of $0.1\%$ were calculated. Finally, the false reject rate (\frr) was estimated at a fixed \fmr\ of $0.1\%$. The reader is referred to the ISO/IEC 19795-1 standard for the definitions of the used performance metrics \cite{ISO-IEC-19795-1-060401}.

\begin{table}[!htb]
    \centering
    \caption{Number of biometric comparisons for each of the considered scenarios and systems.}
    \begin{tabular}{@{}lll@{}} \toprule  \textbf{Scenario} & \multicolumn{1}{c}{\textbf{\arcface}}  & \multicolumn{1}{c}{\textbf{\cots}} \\ \midrule 
    $\pmb{\pazocal{G}_{\pazocal{T}\pazocal{P}}}$ & 460 & 475 \\
    $\pmb{\pazocal{G}_{\pazocal{T}\pazocal{P}_{small}}}$ & 155 & 154 \\
    $\pmb{\pazocal{G}_{\pazocal{T}\pazocal{P}_{medium}}}$ & 191 & 191 \\
    $\pmb{\pazocal{G}_{\pazocal{T}\pazocal{P}_{large}}}$ & 114 & 130 \\
    $\g$ & 4,075 & 4,076 \\
    $\ig$ & 561,804 & 561,804 \\
    \bottomrule
    \end{tabular}
    \label{tab:comparisons_overview}
\end{table}

%% file: sections/results.tex
\section{Results}
\label{sec:results}
The following subsections report the results obtained for the experiments (section \ref{sec:experiments}) on face detection (section \ref{sec:exp_fd}), face quality (section \ref{sec:exp_fq}), as well as feature extraction and comparison (section \ref{sec:exp_fc}).

\subsection{Face Detection}\label{sec:exp_fd}
\begin{figure*}[htb]
\begin{subfigure}[t]{0.48\columnwidth}
  \includegraphics[width=\columnwidth]{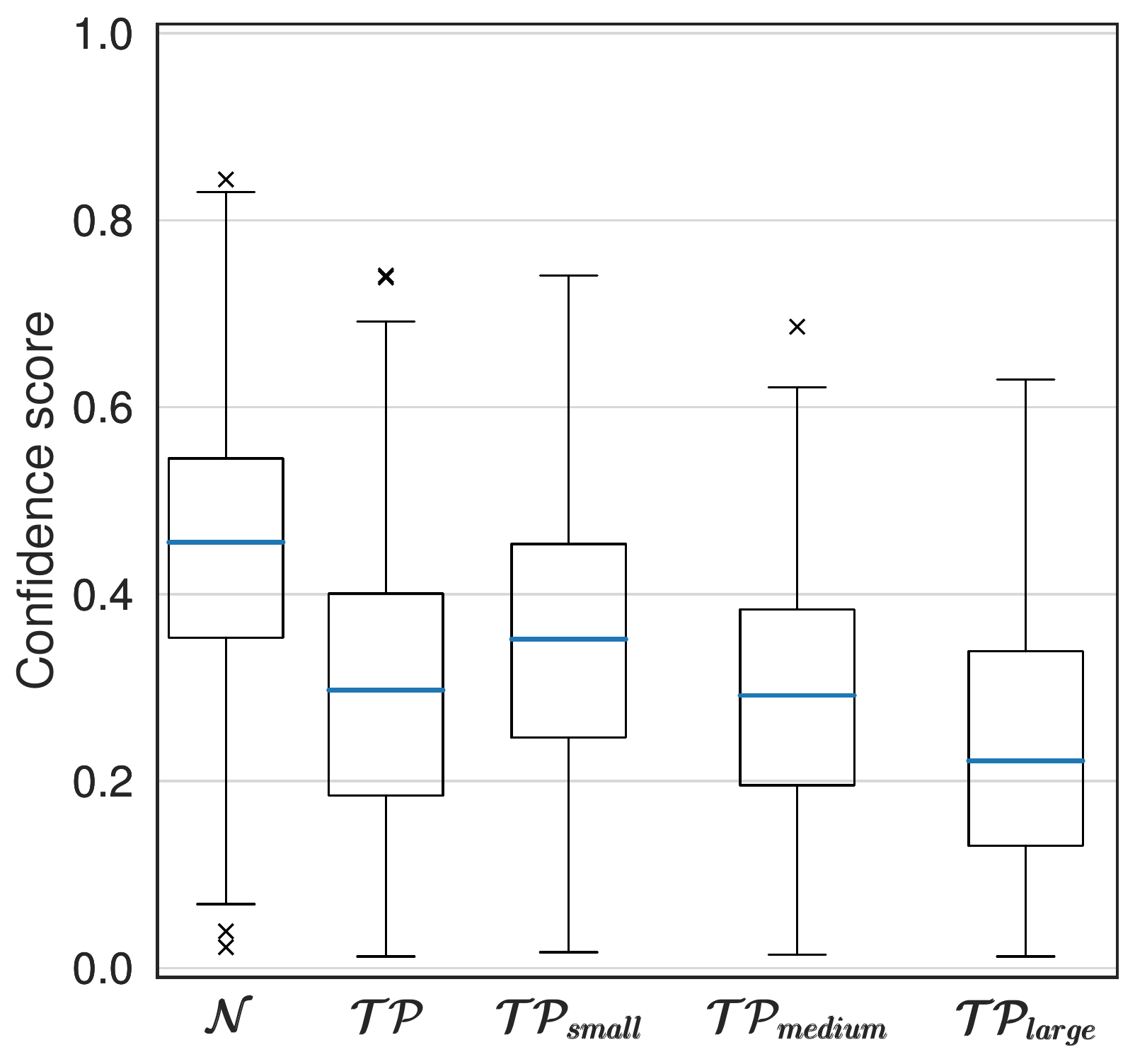}
  \caption{\dlib}
\end{subfigure}\quad %
\begin{subfigure}[t]{0.48\columnwidth}
  \includegraphics[width=\columnwidth]{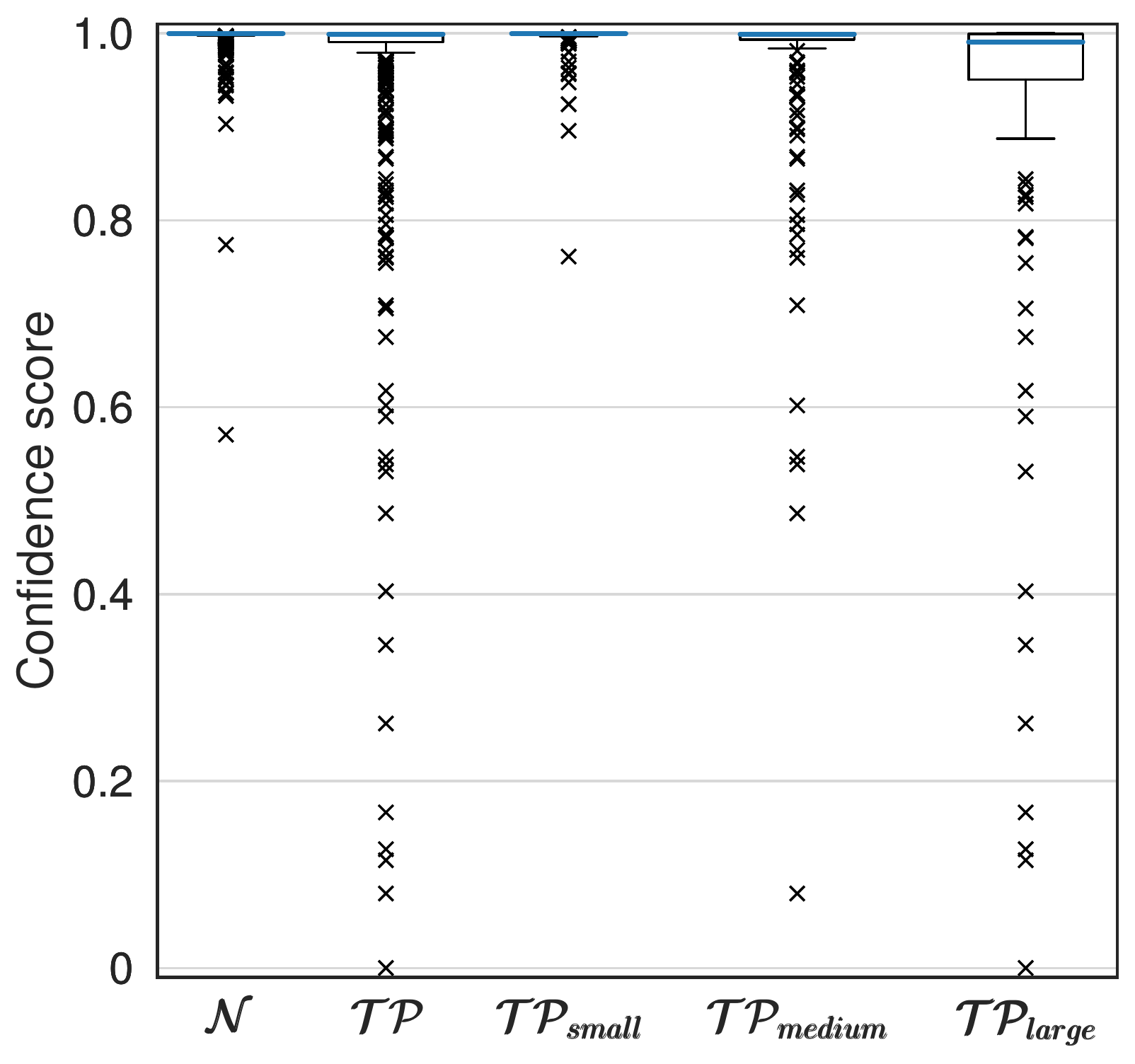}
  \caption{\mtcnn}
\end{subfigure}\quad %
\centering
\begin{subfigure}[t]{0.48\columnwidth}
  \includegraphics[width=\columnwidth]{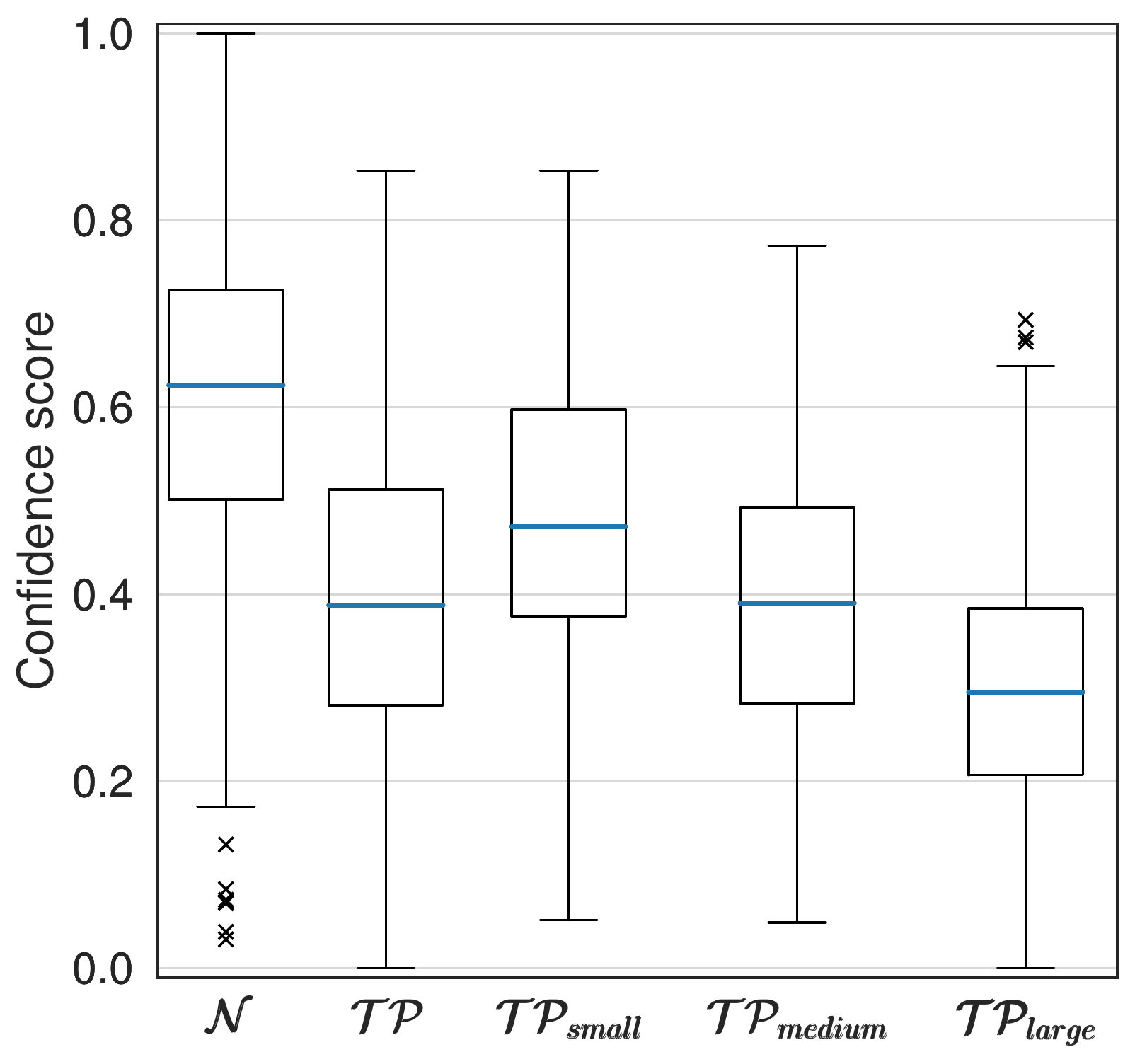}
  \caption{\cots}
  \label{fig:finding_placement_ex_b}
\end{subfigure}\quad %
\caption{Boxplots of face detection confidence scores.}
\label{fig:boxplot_detection}
\end{figure*}

\begin{figure*}[!htb]
\centering
\begin{subfigure}[t]{0.48\columnwidth}
  \includegraphics[width=\columnwidth]{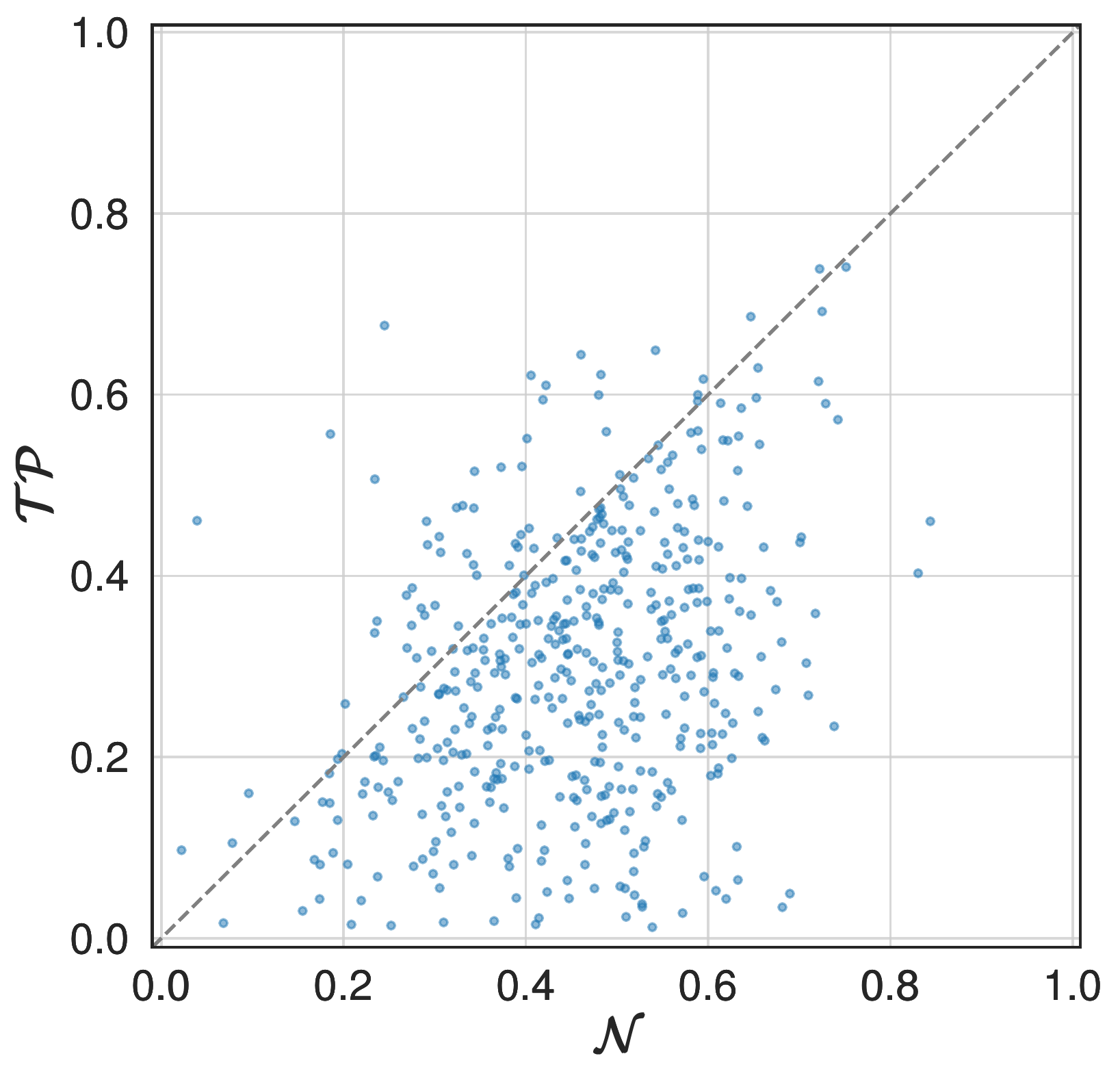}
  \caption{\dlib}
\end{subfigure}\quad %
\begin{subfigure}[t]{0.48\columnwidth}
  \includegraphics[width=\columnwidth]{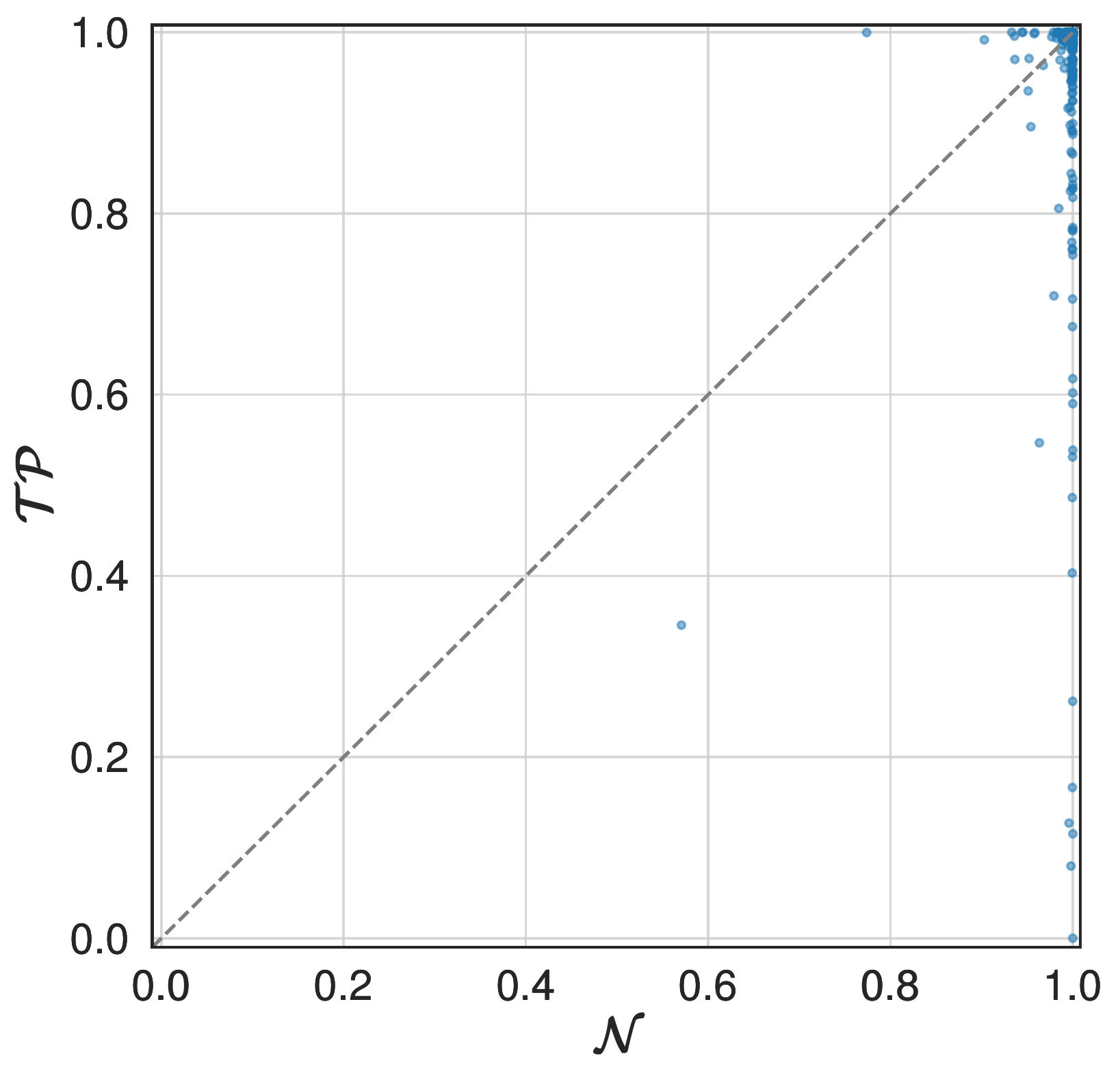}
  \caption{\mtcnn}
\end{subfigure}\quad %
\centering
\begin{subfigure}[t]{0.48\columnwidth}
  \includegraphics[width=\columnwidth]{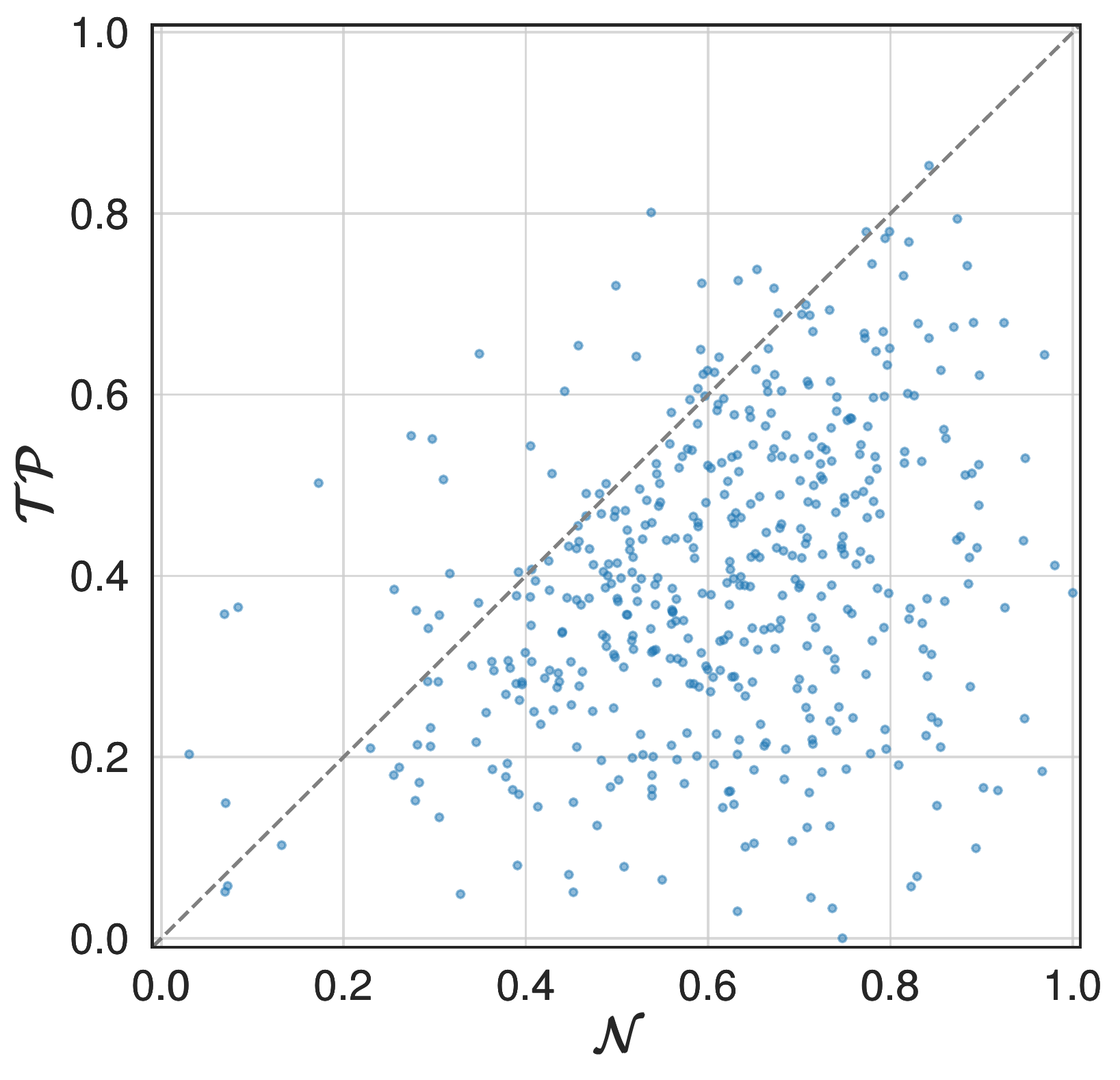}
  \caption{\cots}
\end{subfigure}\quad %
\caption{Comparison of face detection confidence scores for images with and without facial tattoos or paintings.}
\label{fig:detection_scatter}
\end{figure*}

The scatter plots in figure \ref{fig:detection_scatter} compare the detection confidence scores obtained on the image pairs in the assembled database and illustrate that images with tattoos and paintings achieve comparably lower detection scores than the facial images without any manipulations for all three algorithms. These findings are further observed in figure \ref{fig:boxplot_detection} which shows boxplots for the detection confidence scores for different subsets of the assembled database where it can also be observed that it becomes more difficult for algorithms to detect faces when the coverage of tattoos and paintings in a face region is increased. Table \ref{tab:det_scores_dlib_mtcnn_COTS} shows the mean ($\mu$) and standard deviation ($\sigma$) for the obtained detection confidence scores as well as the detection errors (\de) for different subsets of the database. The results show that facial tattoos and paintings decrease the detection confidence scores. Further, they cause significant increases in the percentage of times where the three algorithms fails to detect any faces, \textit{i.e.} all three algorithms have a \de\ score above $10\%$ when a large area of the face has been manipulated by tattoos or paintings.

\begin{table*}[!htb]
    \centering
    \caption{Descriptive statistics of obtained confidence scores and error rates for face detection.}
    \begin{tabular}{@{\extracolsep{2pt}}llllllllll@{}} \toprule 
    \multirow{2}{*}{} &
      \multicolumn{3}{c}{\textbf{\dlib}}  &
      \multicolumn{3}{c}{\textbf{\mtcnn}} &
      \multicolumn{3}{c}{\textbf{\cots}} \\ \cmidrule{2-4} \cmidrule{5-7} \cmidrule{8-10}
    \textbf{Subset} & $\mu $& $\sigma$ & \de $\%$ & $\mu $& $\sigma$ & \de \% & $\mu $& $\sigma$ & \de \%\\ \midrule
    $\n$ & $0.45$ & $0.14$ & $0.60$ & $1.00$ & $0.02$ & $1.00$ & $0.61$ & $0.17$ & $0.60$ \\
    $\tp$ & $0.30$ & $0.15$ & $7.00$ & $0.96$ & $0.13$ & $7.00$ & $0.39$ & $0.16$ & $5.80$ \\
    $\tps$ & $0.36$ & $0.15$ & $1.28$ & $0.99$ & $0.02$ & $0.00$ & $0.48$ & $0.16$ & $0.64$ \\
    $\tpm$& $0.29$ & $0.14$ & $6.60$ & $0.97$ & $0.10$ & $2.54$ & $0.39$ & $0.15$ & $4.06$ \\
    $\tpl$ & $0.23$ & $0.14$ & $13.61$ & $0.91$ & $0.20$ & $20.41$ & $0.31$ & $0.14$ & $13.61$ \\
    \bottomrule
    \end{tabular}
    \label{tab:det_scores_dlib_mtcnn_COTS}
\end{table*}

\begin{figure*}[!htb]
\begin{subfigure}[t]{0.48\columnwidth}
  \includegraphics[width=\columnwidth]{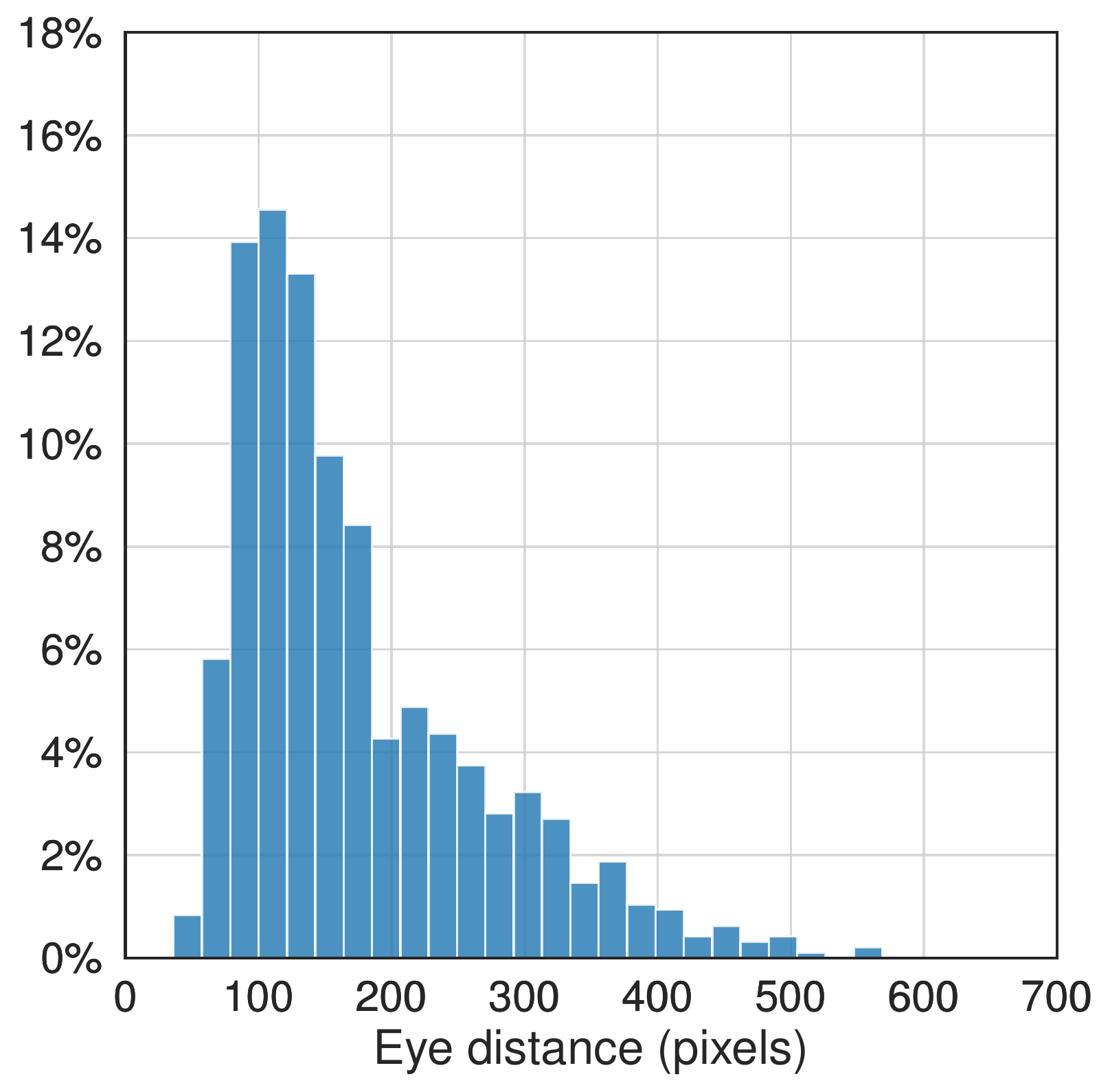}
  \caption{\dlib}
\end{subfigure}\quad %
\begin{subfigure}[t]{0.48\columnwidth}
  \includegraphics[width=\columnwidth]{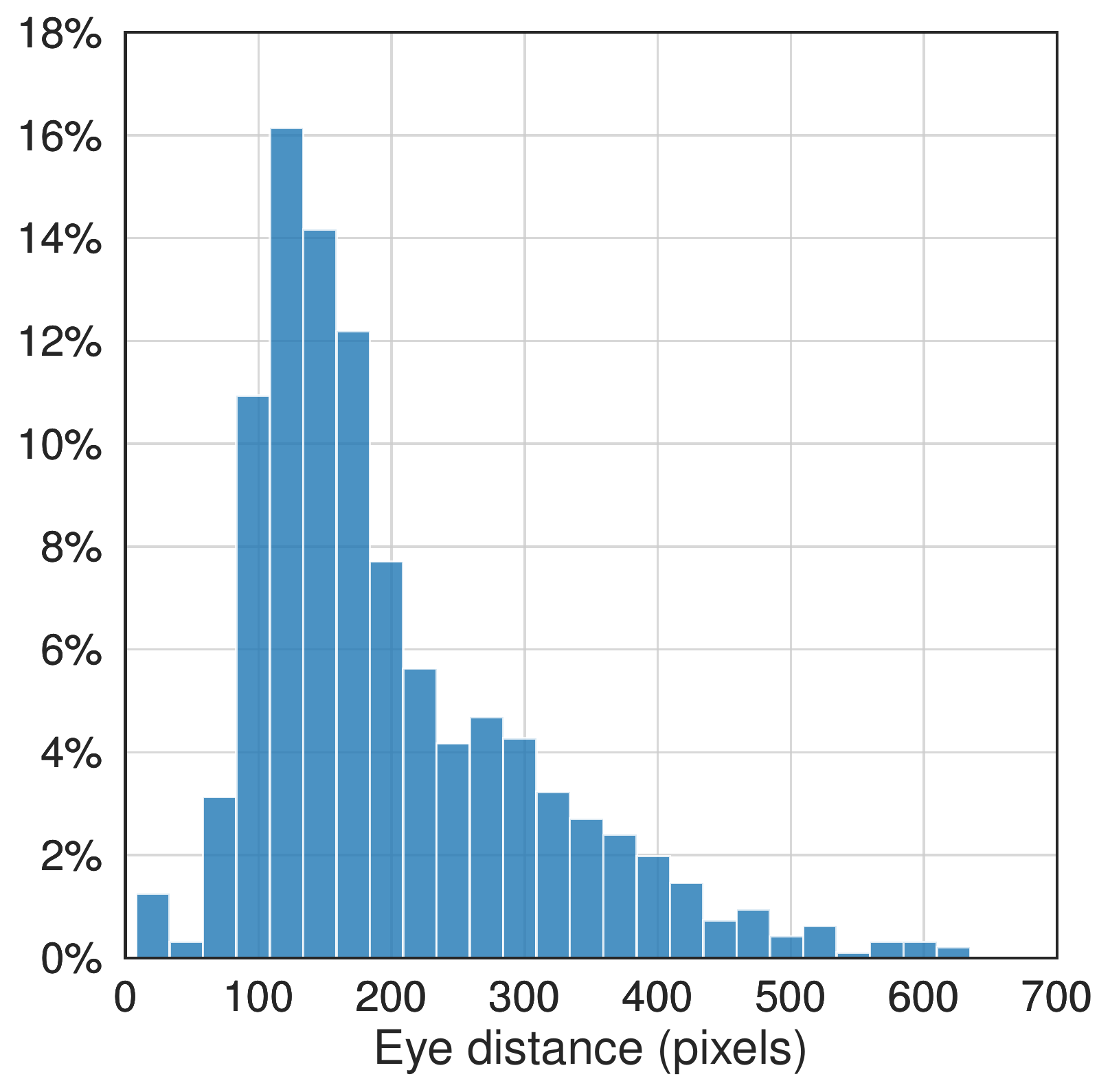}
  \caption{\mtcnn}
\end{subfigure}\quad %
\centering
\begin{subfigure}[t]{0.48\columnwidth}
  \includegraphics[width=\columnwidth]{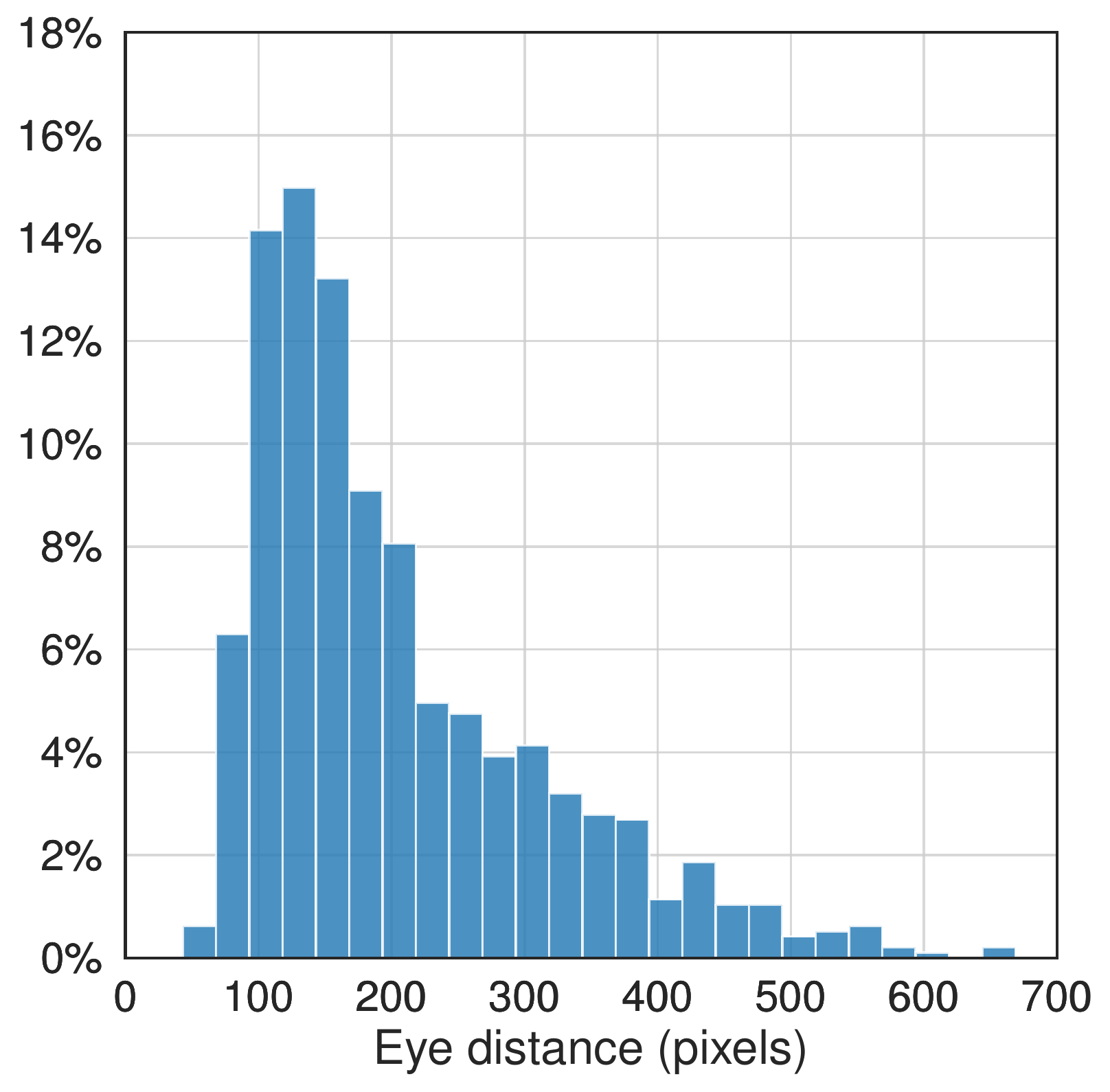}
  \caption{\cots}
\end{subfigure}\quad %
\caption{Eye distances of detected faces. }
\label{fig:eyedistance_detection}
\end{figure*}

\begin{figure}[ht]
\centering
\begin{subfigure}[t]{0.2304\columnwidth}
    \centering
  \includegraphics[width=\columnwidth]{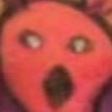}
\end{subfigure} %
\begin{subfigure}[t]{0.2304\columnwidth}
    \centering
  \includegraphics[width=\columnwidth]{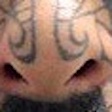}
\end{subfigure} %
\begin{subfigure}[t]{0.2304\columnwidth}
    \centering
  \includegraphics[width=\columnwidth]{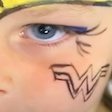}
\end{subfigure} %
\begin{subfigure}[t]{0.2304\columnwidth}
    \centering
  \includegraphics[width=\columnwidth]{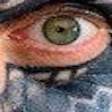}
\end{subfigure} %
\caption{Examples where the detection with the highest confidence score was not actually a face.} 
\label{fig:incorrect_detections}
\end{figure}

\begin{figure}[ht]
\centering
\begin{subfigure}[t]{0.2304\columnwidth}
    \centering
  \includegraphics[width=\columnwidth]{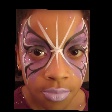}
  \caption{\dlib}
\end{subfigure}\quad %
\begin{subfigure}[t]{0.2304\columnwidth}
    \centering
  \includegraphics[width=\columnwidth]{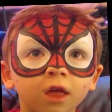}
  \caption{\mtcnn}
\end{subfigure}\quad %
\begin{subfigure}[t]{0.2304\columnwidth}
    \centering
  \includegraphics[width=\columnwidth]{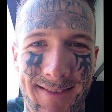}
  \caption{\cots}
\end{subfigure}\quad %
\caption{Examples where the considered systems failed to detect any faces.}
\label{fig:mtcnn_dlib_fail}
\end{figure}

Not all detections where an algorithm produced the highest confidence value correspond to a correctly detected face. Figure \ref{fig:eyedistance_detection} illustrates the distances between the eyes of the detected faces for each algorithm. While most of the eye distances are around 100 pixels or above several outliers are also shown, which contain cases where faces have been incorrectly detected. Figure \ref{fig:incorrect_detections} shows examples of incorrect detections. Additionally, examples of facial images where zero faces were detected by one of the algorithms are shown in figure \ref{fig:mtcnn_dlib_fail}. An analysis of the failure cases reveals that on many images where the algorithms failed to detect any faces, manipulations have occurred near the periocular region or a large area of the face was covered in tattoos or paintings including areas around the mouth, nose and eyes. However, it is not always the case that the algorithms fail to detect a face despite manipulations near the aforementioned areas.

\raggedcolumns
\subsection{Face Quality Estimation}\label{sec:exp_fq}
\begin{figure}[!htb]
\begin{subfigure}[t]{0.48\columnwidth}
    \centering
  \includegraphics[width=\columnwidth]{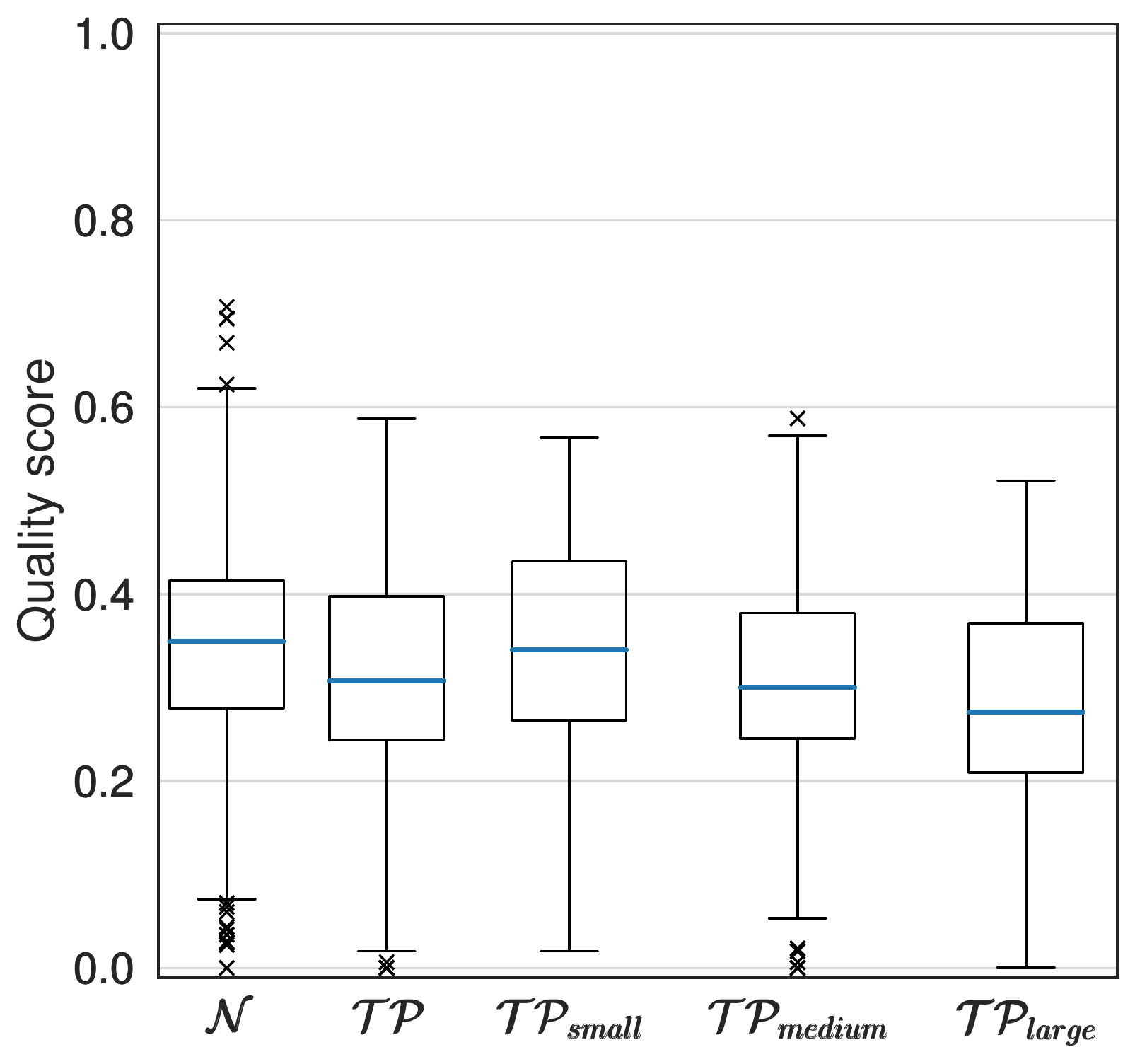}
  \caption{\faceqnet}
\end{subfigure}\quad %
\begin{subfigure}[t]{0.48\columnwidth}
    \centering
  \includegraphics[width=\columnwidth]{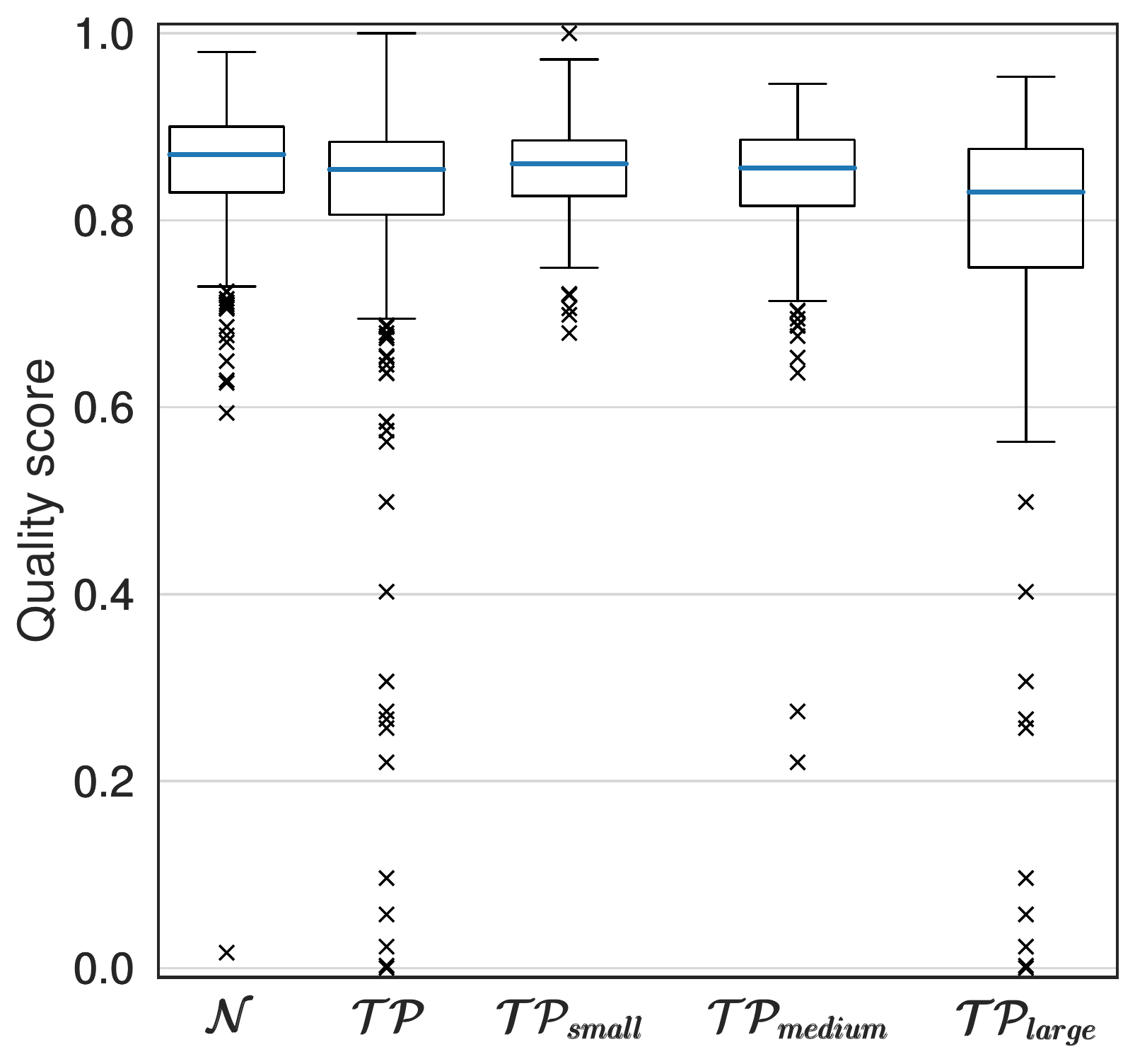}
  \caption{\serfiq}
\end{subfigure}\quad %
\caption{Boxplots of the estimated quality scores.}
\label{fig:boxplot_quality}
\end{figure}

\begin{figure}[!htb]
\begin{subfigure}[t]{0.48\columnwidth}
    \centering
  \includegraphics[width=\columnwidth]{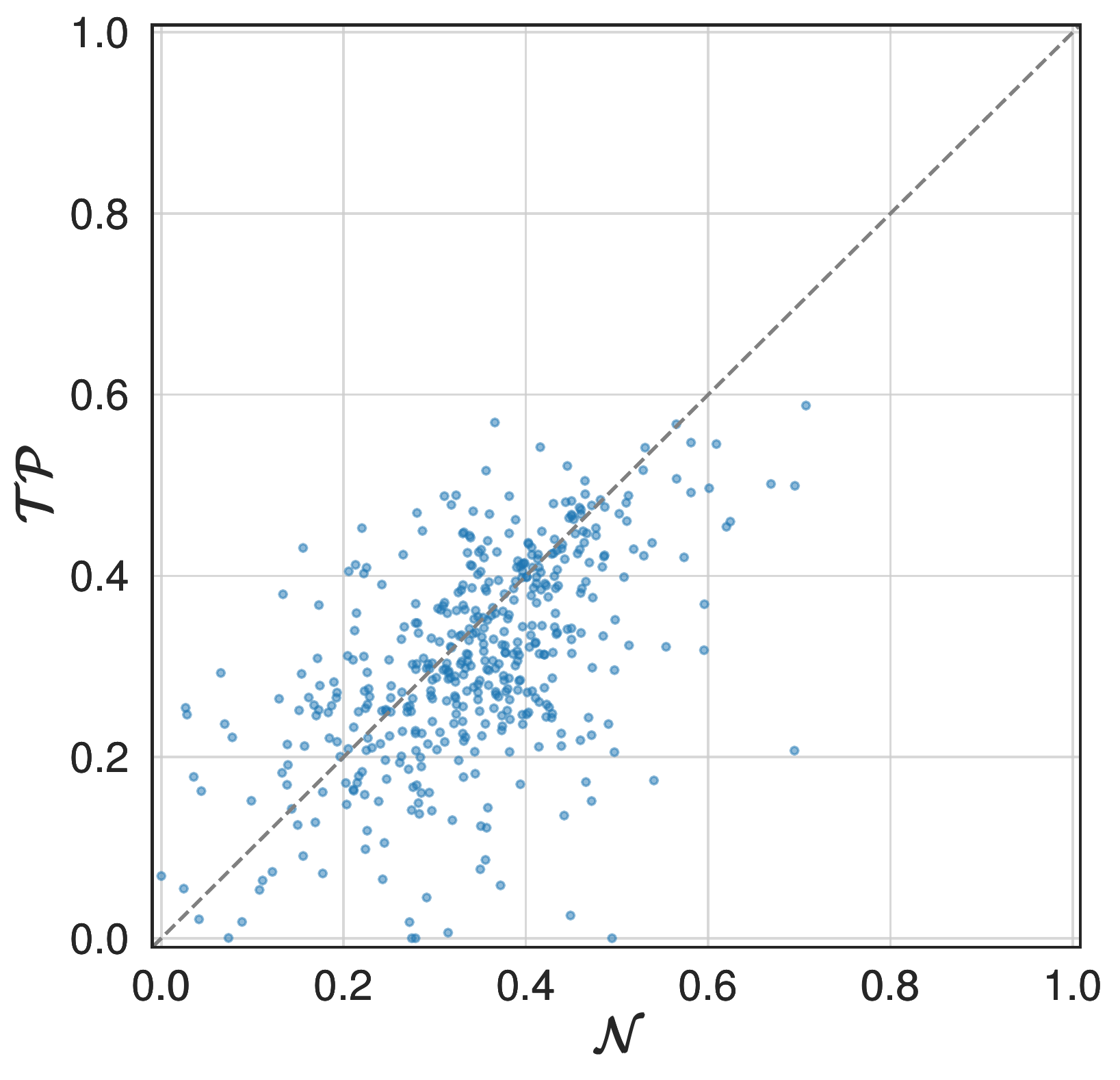}
  \caption{\faceqnet}
\end{subfigure}\quad %
\begin{subfigure}[t]{0.48\columnwidth}
    \centering
  \includegraphics[width=\columnwidth]{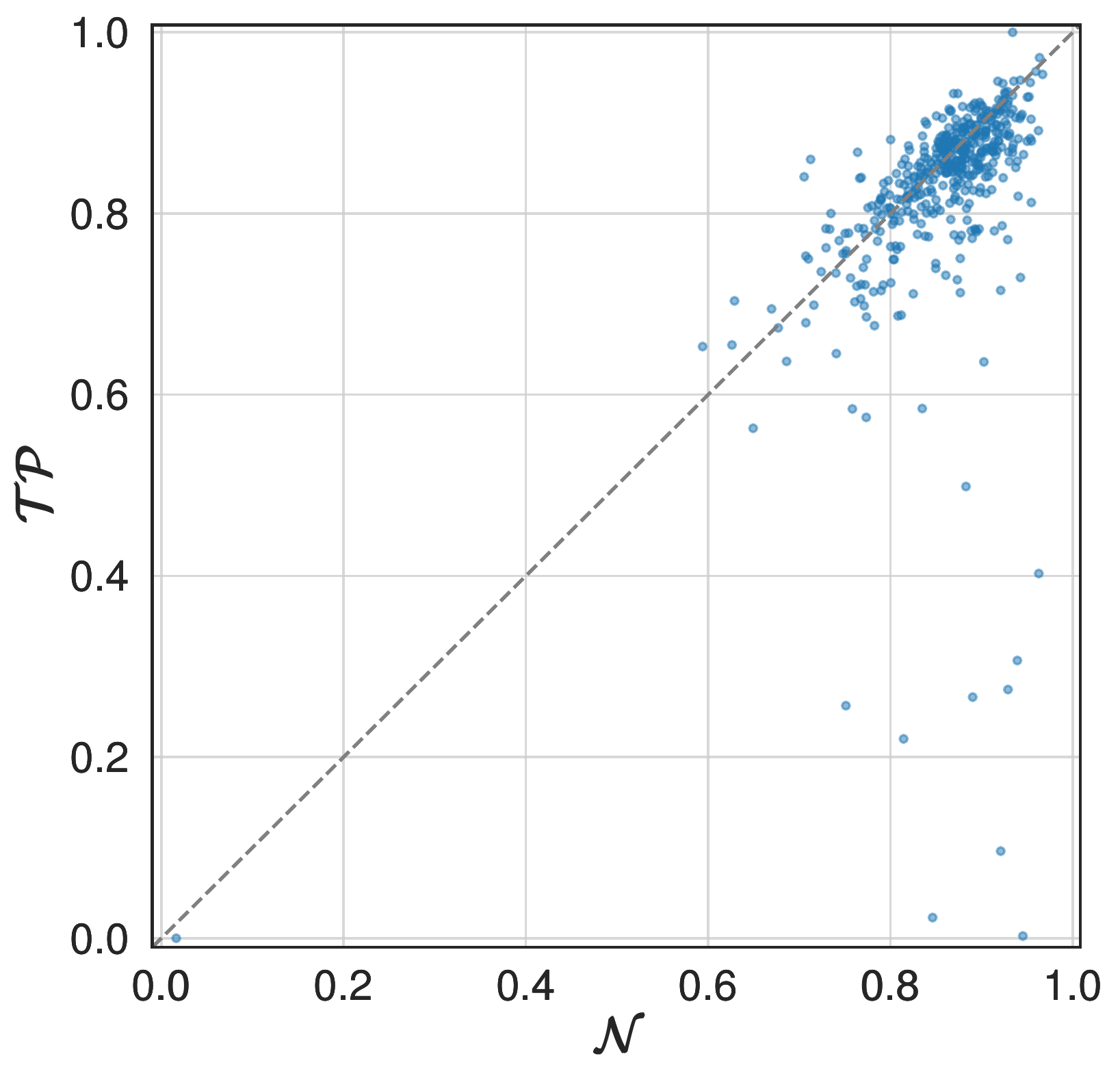}
  \caption{\serfiq}
\end{subfigure}\quad %
\caption{Comparison of quality scores for images with and without facial tattoos or paintings.}
\label{fig:scatter_quality}
\end{figure}

From the comparisons illustrated in figure \ref{fig:scatter_quality}, it can be noted that facial tattoos and paintings have some impact on the quality estimation scores of \serfiq and \faceqnet. From figure \ref{fig:boxplot_quality} and table \ref{tab:table_quality}, it can be seen that especially medium and large coverage of tattoos and paintings affect the quality estimation of \faceqnet and \serfiq, whereas smaller manipulations have an insignificant impact. \par
\begin{figure}[!htbp]
    \centering
\begin{subfigure}[t]{0.2304\columnwidth}
    \centering
  \includegraphics[width=\columnwidth]{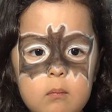}
\end{subfigure}
\begin{subfigure}[t]{0.2304\columnwidth}
    \centering
  \includegraphics[width=\columnwidth]{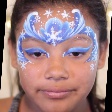}
\end{subfigure}
\begin{subfigure}[t]{0.2304\columnwidth}
    \centering
  \includegraphics[width=\columnwidth]{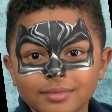}
\end{subfigure}
\begin{subfigure}[t]{0.2304\columnwidth}
    \centering
  \includegraphics[width=\columnwidth]{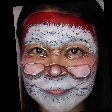}
\end{subfigure} %
\caption{Facial images which obtained low quality scores using \faceqnet.}
\label{fig:faceqnet_badescores}
\end{figure}

 Analysis of the $\tp$ images which received the lowest quality scores reveal that especially manipulations around the periocular region contribute to a low quality score; examples are shown in figure \ref{fig:faceqnet_badescores}. Furthermore, as \mtcnn was used for both \serfiq and \faceqnet (see section \ref{subsec:exp_face_quality}), a low quality score was sometimes obtained when a face was not correctly detected (\textit{e.g.} as in figure \ref{fig:incorrect_detections}).

\begin{table}[!htbp]
    \centering
\caption{Descriptive statistics of the estimated quality scores.}
    \begin{tabular}{@{\extracolsep{2pt}}llllllllll@{}} \toprule 
     &
      \multicolumn{2}{c}{\textbf{\faceqnet}}  &
      \multicolumn{2}{c}{\textbf{\serfiq}} \\ \cmidrule{2-3}  \cmidrule{4-5}
    \textbf{Subset} & $\mu $& $\sigma$ & $\mu $& $\sigma$ \\ \midrule
    $\n$ &$0.3417$ & $0.1148$  & $0.8586$  & $0.0707$ \\
    $\tp$ & $0.3091$ & $0.1132$ & $0.8252$ &  $0.1233$ \\
    $\tps$ & $0.3431$ & $0.1113$ & $0.8545$ & $0.0533$  \\
    $\tpm$ & $0.2989$ & $0.1103$  & $0.8381$ & $0.0854$ \\
    $\tpl$ & $0.2806$ & $0.1102$ & $0.7651$ & $0.1999$ \\
    \bottomrule
    \end{tabular}
\label{tab:table_quality}
\end{table}

\subsection{Feature Extraction and Comparison}\label{sec:exp_fc}
\begin{figure}[ht]
\centering
    \centering
  \includegraphics[width=0.2304\columnwidth]{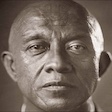}
   \includegraphics[width=0.2304\columnwidth]{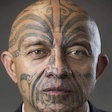}
    \caption{Example of quality differences between two images in an image pair. \arcface correctly enrols the facial image with tattoos (right), but fails for the facial image without tattoos (left).}
\label{fig:quality_diff_example}
\end{figure}

\begin{figure}[!htb]
\begin{subfigure}[t]{0.48\columnwidth}
    \centering
  \includegraphics[width=\columnwidth]{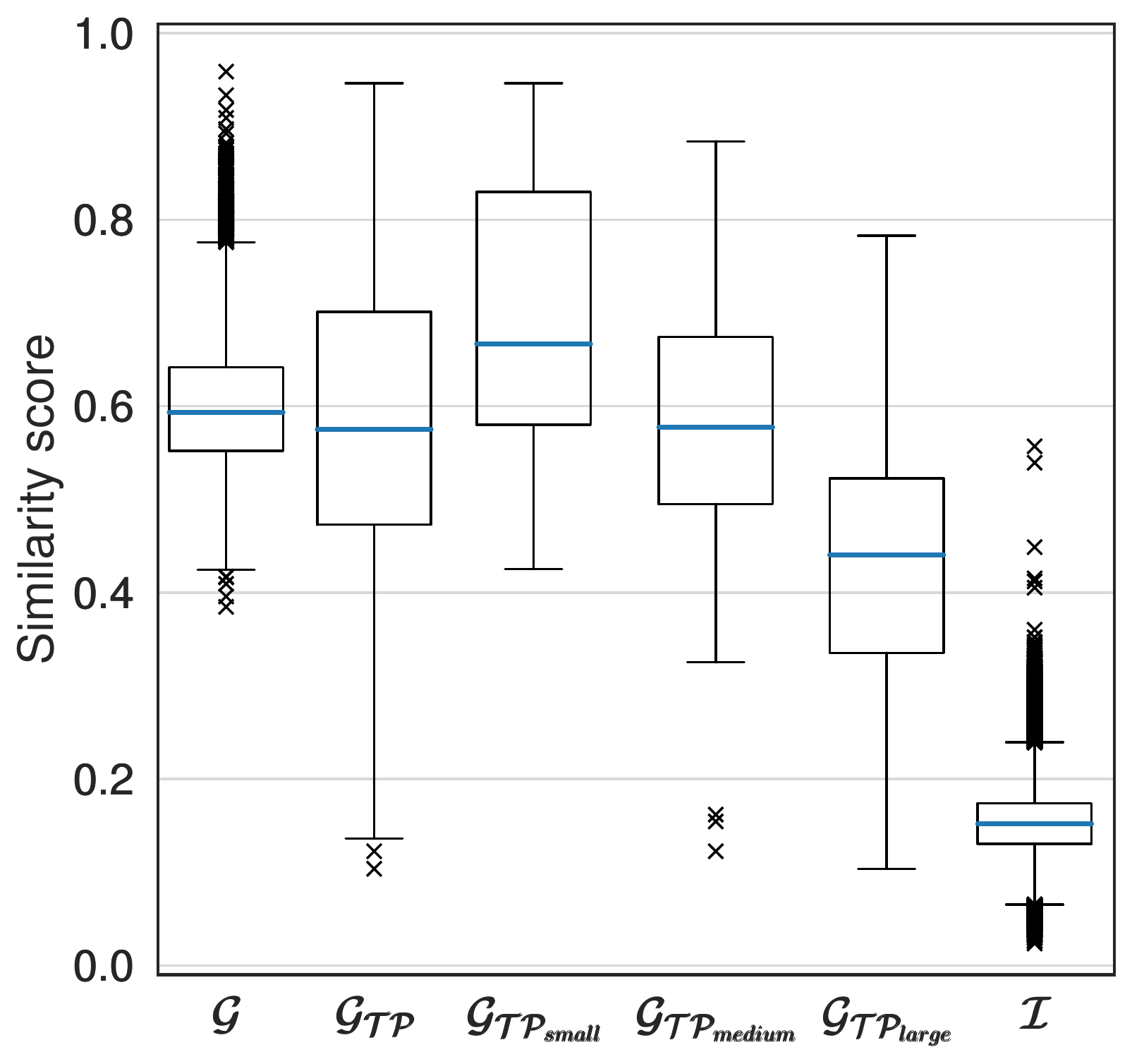}
  \caption{\arcface}
\end{subfigure}\quad %
\begin{subfigure}[t]{0.48\columnwidth}
    \centering
  \includegraphics[width=\columnwidth]{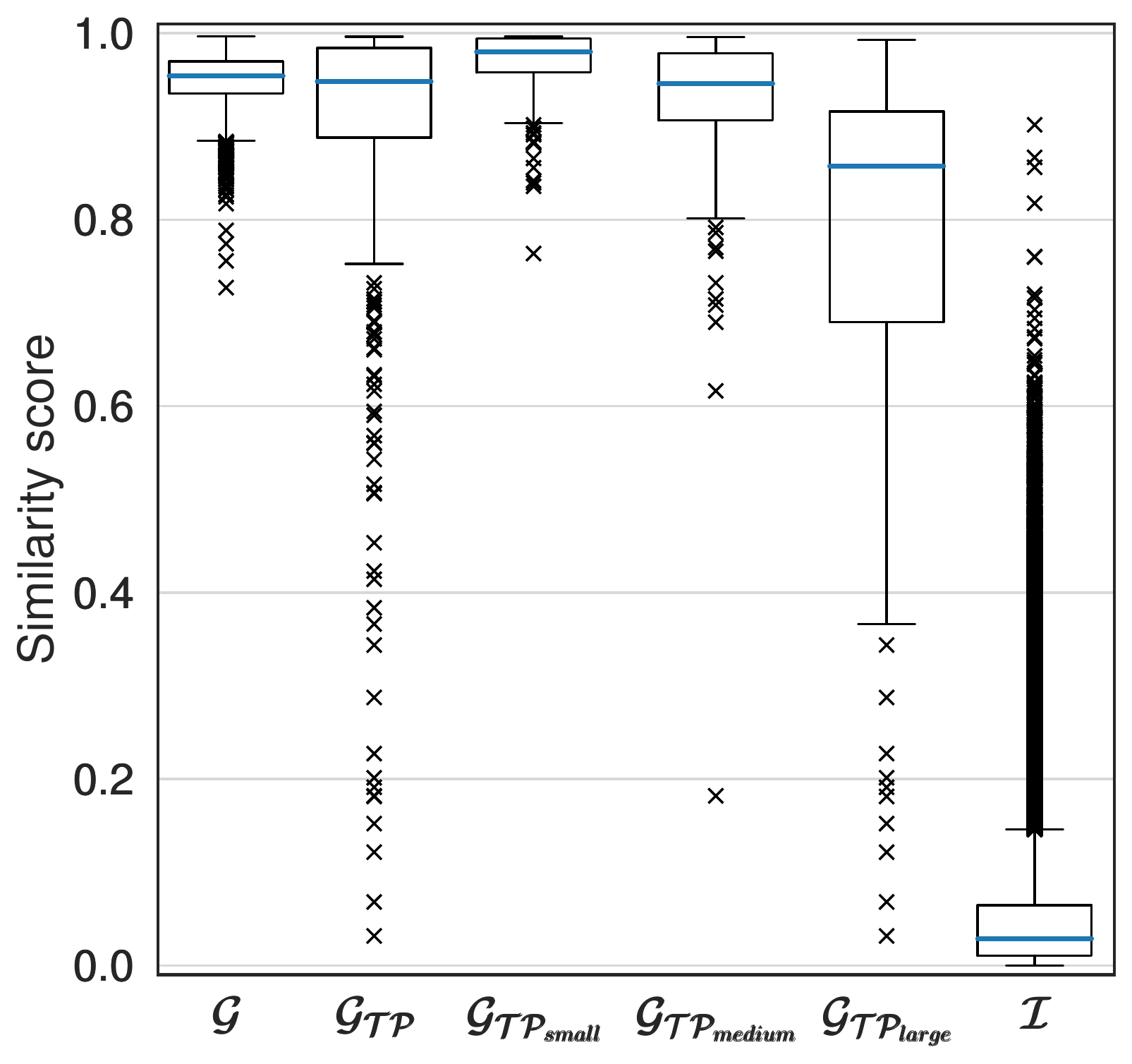}
  \caption{\cots}
\end{subfigure}\quad %
\caption{Boxplots of comparison scores.}
\label{fig:boxplots_verification}
\end{figure}

The boxplots in figure \ref{fig:boxplots_verification} shows the comparison scores obtained for \cots and \arcface on the genuine and imposter comparisons explained in section \ref{subsec:face_verification}. The boxplots show that especially large tattoos and paintings have an impact on the recognition performance of the tested systems, whereas smaller manipulations have no significant impact. From the statistical results of the obtained comparison scores in table \ref{tab:statistical_data_verification}, it can be observed that the comparison scores are significantly reduced for both \arcface and \cots in cases where a large area of a face is covered with tattoos or paintings. Furthermore, a relatively high standard deviation of comparison scores for large facial tattoos and paintings can be seen for the \cots system (${\sim}0.226$). These observations indicate that especially some extreme cases contribute to the performance degradation observed in the comparison module of the tested commercial system. The performance scores reported in table \ref{tab:biometric_performance_arcface} for \arcface, show that facial tattoos and paintings both influence the system's ability to extract features and its ability to verify the identity of individuals. For $\pmb{\pazocal{G}_{\pazocal{T}\pazocal{P}_{large}}} \times \pmb{\pazocal{I}}$ on \arcface, a high \frr\ (${\sim}30.9\%$) was observed at a \fmr\ of $0.1\%$. This indicates that if such a system were to be deployed, many manual interventions might be required for individuals who have large facial tattoos or paintings. Notably, high equal error rate was observed for $\pmb{\pazocal{G}_{\pazocal{T}\pazocal{P}_{medium}}}$ and $\pmb{\pazocal{G}_{\pazocal{T}\pazocal{P}_{large}}}$ compared to the impostor and genuine comparisons where no facial image have tattoos or paintings ($\pmb{\pazocal{G}} \times \pmb{\pazocal{I}}$). Similar tendencies can be observed for \cots in table \ref{tab:biometric_performance_COTS} where it is observed that both medium and especially large manipulations have an impact on its biometric performance. From table \ref{tab:failure_to_extract}, it can be observed that both systems achieve high enrolment errors. Even for the best performing system (\cots), more than $10\%$ of the facial images where a large area of the face are covered with tattoos or paintings, cannot be enrolled in the system. In table \ref{tab:failure_to_extract}, it can also be observed that the \fte\ for $\tps$ is $0$, but the \fte\ for $\n$ is higher for both \arcface and \cots. The reason for this observation is that the facial images in the collected database have been assembled from online sources; hence, differences in acquisition conditions and quality between two facial images in an image pair can occur. An example is shown in figure \ref{fig:quality_diff_example}, where the differences in acquisition conditions between the two facial images cause \arcface to correctly enrol the facial image containing tattoos whereas it fails to enrol the facial image without any tattoos. The \fte\ of $0$ observed for $\tps$, therefore, means that all the facial images in this subset are of sufficient quality and that small coverage of tattoos and paintings on a face has not affected the tested systems' ability to enrol them. \par
\begin{figure}[!htb]
\centering
\begin{subfigure}[t]{.48\columnwidth}
\centering
\vspace{0.1pt}
\includegraphics[width=.48\columnwidth]{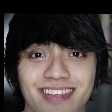}
\includegraphics[width=.48\columnwidth]{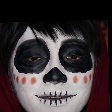}
\caption{\arcface}
\end{subfigure}\quad
\begin{subfigure}[t]{.48\columnwidth}
\centering
\vspace{0.1pt}
\includegraphics[width=.48\columnwidth]{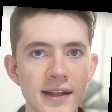}
\includegraphics[width=.48\columnwidth]{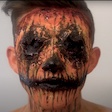}
\caption{\cots}
\end{subfigure}
\caption{Examples of image pairs where a low comparison score was achieved.}
\label{fig:low_comparison_scores}
\end{figure}

 An inspection of the $\tp$ images which contributed to the lowest comparison scores achieved for the  $\pmb{\pazocal{G}_{\pazocal{T}\pazocal{P}}} \times \pmb{\pazocal{I}}$ comparisons of both \arcface and \cots reveals that low comparison scores were especially achieved when the prepossessing step failed to accurately detect the face (\textit{e.g.} as in figure \ref{fig:incorrect_detections}) or in extreme cases where most of the face is covered with tattoos or paintings as illustrated in figure \ref{fig:low_comparison_scores}. Similar, to what has been observed in other works \textit{e.g.} \cite{Singh-RecognizingDisguisedFacesInTheWild-TBIOM-2019}, there is a tendency that face recognition systems are sensitive to manipulations near the periocular region. However, more control over which manipulations occur in what regions of the face is needed to conclude this. To this end, synthetic data could be generated which is discussed more in section \ref{sec:discussion}.

\begin{table}[!htbp]
    \centering
\caption{Descriptive statistics of the comparison scores.}
    \begin{tabular}{@{\extracolsep{2pt}}lllll@{}} \toprule 
       &
      \multicolumn{2}{c}{\textbf{\arcface}}  &
      \multicolumn{2}{c}{\textbf{\cots}} \\  \cmidrule{2-3} \cmidrule{4-5}
    \textbf{Scenario} & $\mu $& $\sigma$ & $\mu $& $\sigma$ \\ \hline
    $\pmb{\pazocal{G}_{\pazocal{T}\pazocal{P}}}$ & $0.5840$ & $0.1700$ & $0.8974$ & $0.1531$  \\
    $\pmb{\pazocal{G}_{\pazocal{T}\pazocal{P}_{small}}}$ & $0.6920$ & $0.1378$ & $0.9661$ & $ 0.0397$  \\
    $\pmb{\pazocal{G}_{\pazocal{T}\pazocal{P}_{medium}}}$ & $0.5868$ & $0.1347$ & $0.9275$ & $ 0.0840$  \\
    $\pmb{\pazocal{G}_{\pazocal{T}\pazocal{P}_{large}}}$ & $0.4325$ & $0.1476$ &$0.7718$ & $0.2257$ \\
    $\pmb{\pazocal{I}}$ & $0.1530$ & $0.0330$ & $0.0478$ & $ 0.0559$  \\
    $\pmb{\pazocal{G}}$ & $0.6058$ & $0.0823$ & $0.9509$ & $ 0.0289$  \\
    \bottomrule
    \end{tabular}
\label{tab:statistical_data_verification}
\end{table}

\begin{table}[!htbp]
    \centering
\caption{Failure to enrol rate for \arcface and \cots on different subsets of the assembled database. No failures occurred during enrolment for the reference images used from the FERET and FRGCv2 databases.}
   \begin{tabular}{@{\extracolsep{2pt}}lll@{}} \toprule 
       &
      \multicolumn{1}{c}{\textbf{\arcface}}  &
      \multicolumn{1}{c}{\textbf{\cots}} \\  \cmidrule{2-2} \cmidrule{3-3}
    \textbf{Subset} & \fte $\%$ & \fte $\%$ \\ \hline
    $\n$  &  $1.0$ & $0.6$ \\
    $\tp$ & $7.0$ & $4.4$  \\
    $\tps$ & $0.0$ & $0.0$ \\
    $\tpm$ & $2.5$ & $3.0$  \\
    $\tpl$ & $20.4$ & $10.9$   \\
    \bottomrule
    \end{tabular}
\label{tab:failure_to_extract}
\end{table}

\begin{table*}[!htb]
    \centering
\caption{Biometric performance results for \arcface and \cots. \fnmr\ and \frr\ are calculated at \fmr\ = 0.1\%.}
   \begin{tabular}{@{\extracolsep{2pt}}lllllll@{}} \toprule 
       & \multicolumn{3}{c}{\textbf{\arcface}}  &
      \multicolumn{3}{c}{\textbf{\cots}} \\  \cmidrule{2-4} \cmidrule{5-7}
\textbf{Comparison} & \eer $\%$ & \fnmr $\%$ & \frr  $\%$ & \eer $\%$ & \fnmr $\%$ & \frr $\%$  \\ \midrule

        $\pmb{\pazocal{G}_{\pazocal{T}\pazocal{P}}} \times \pmb{\pazocal{I}}$ & $2.8261$ & $3.9130$ & $10.6391$ & $1.6985$ & $2.9474$ & $7.2177$ \\

        $\pmb{\pazocal{G}_{\pazocal{T}\pazocal{P}_{small}}} \times \pmb{\pazocal{I}} $ & $0.0003$ & $0.0000$ & $0.0000$ & $0.0004$ & $0.0000$ & $0.0000$ \\
        
        $\pmb{\pazocal{G}_{\pazocal{T}\pazocal{P}_{medium}}} \times  \pmb{\pazocal{I}}$ & $1.5707$ & $1.5707$ & $4.0689$ & $ 0.5235$ & $0.5236$ & $3.5533$ \\
        
         $\pmb{\pazocal{G}_{\pazocal{T}\pazocal{P}_{large}}} \times  \pmb{\pazocal{I}} $ & $8.7719$ & $13.1579$ & $30.8808$ & $3.2824$ & $10.0000$ & $19.7959$\\
         
        $\pmb{\pazocal{G}} \times \pmb{\pazocal{I}}$ & $0.0005$ & $0.0000$ & $0.0000$ & $0.0005$ & $0.0000$ & $0.0000$ \\
        \bottomrule
    \end{tabular}
\label{tab:biometric_performance_arcface}
\label{tab:biometric_performance_COTS}
\end{table*}

%% file: sections/discussion.tex
\section{Discussion}
\label{sec:discussion}
From the results in section \ref{sec:results}, it can be observed that a large coverage of facial tattoos or paintings in a face has a significant impact on the tested algorithms. More specifically, the following observations can be made from the results:

\begin{description}
\item[\textbf{Face detection} ] The negative impact observed when a large area of a face is covered with tattoos or paintings indicate that individuals might successfully apply facial tattoos or paint to obfuscate their identity and avoid being recognised or enrolled by biometric systems. The focus of this work was not on the deliberate use of facial tattoos and paintings to undermine the security of a biometric system. However, it is not unlikely that individuals purposely employ such manipulations to avoid being recognised.
\end{description}

\begin{description}
\item[\textbf{Face quality estimation}] From the results, it can be observed that the performance of \faceqnet and \serfiq decreases as the coverage of facial tattoos and paintings on a face become large. Since face quality estimation measures the utility of a facial image for face recognition \cite{Schlett-FaceImageQualityAssessmentALiteratureSurvey}, the observations are expected since for the comparison and features extraction module of both \arcface and \cots a decrease in biometric performance was observed for facial tattoos and paintings compared to facial images without such manipulations.
\end{description}

\begin{description}
\item[\textbf{Feature extraction and comparison}] In the conducted experiments, we showed that facial tattoos and paintings significantly impact face recognition systems' ability to extract face features and recognise individuals. However, in the experiments, we only considered cases where facial images without manipulations were compared to facial images with either facial tattoos or paintings. It could, however, be the case that the comparison scores are not impacted when both reference and probe image have such manipulations. For instance, in \cite{Eckert-FacialCosmeticsDatabaseAndImpactAnalysisOnAutomaticFaceRecognition} Eckert \textit{et al.} showed that for the conducted experiments better identification rates were observed when both reference and probe image had facial cosmetics compared to when only one of the facial images had such manipulations. Similarly, it could be the case that face recognition systems see facial tattoos and paintings as factors which can distinguish individuals. In fact, tattoos are frequently used by law enforcement officers to identify suspects and victims during criminal investigations \cite{ansi_nist_itl_2011}, \cite{tate_online}.
\end{description}

The images in the assembled database were captured from various online sources and as such images in a pairing may vary in factors such as age, quality, and illumination. Although changes in such factors can influence the biometric performance of a system, the database creation process aimed to minimise those factors. Additionally, significant performance degradation was observed for large manipulations compared to the facial images without any manipulations which indicate that it is, in fact, the facial tattoos and paintings and not other factors which causes the observed degradations. \par The findings in this work indicate a need to design algorithms robust to facial tattoos and paintings. Two approaches for mitigating the impact that facial tattoos and paintings have on face recognition systems can be investigated. One approach is to retrain a CNN on images containing facial images with tattoos and paintings. A similar approach was proposed in \cite{Trigueros-EnchancingConvolutionalNeuralNetworks-2018} for partially occluded faces where the authors proposed training a CNN model using a data set containing synthetically occluded faces and found that it improved robustness to face occlusions. The other approach is to train a classifier capable of removing facial tattoos or paintings from face images. For instance, conditional generative adversarial networks (cGANs) for general purpose image-to-image translation can be used by feeding into the network pairs of images with and without facial manipulations. Doing so will, theoretically, teach the generator network to undo the manipulated part of the image. For both approaches, large datasets of images with and without facial tattoos and paintings are needed, and to this end, synthetic data can be generated and used. Synthetically generating face images with facial tattoos or paintings is a non-trivial task. First, suitable placements of facial tattoos and paintings in a face must be found, and it must be ensured that the manipulations are placed inside the face region and not on top of undesired areas such as glasses or within the eyes, nostrils, or mouth. Secondly, the facial tattoos or paintings must be blended to the face image realistically, which means that the facial tattoos and paintings should be blended to follow the contours of the face. Besides algorithmic changes, political initiatives can be taken and as such a new law can be established, requiring citizens to renew their identification documents after getting facial tattoos. \par Additionally, future work could focus on the impact of demographic factors \cite{Drozdowski-BiasSurvey-TTS-2020} on facial manipulations and a more detailed study to quantify and differentiate between the impacts of facial tattoos and paintings could be conducted. Initial experiments on the assembled database show some differences between tattoos and paintings and some demographic subdivisions of the data. However, as it was shown that the size of the alteration significantly impacts the biometric performance of face recognition systems, more control of the sizes of manipulations is needed before definitive conclusions can be drawn.

%% file: sections/conclusion.tex
\section{Conclusion}
\label{sec:conclusion}
In this work, the impact of facial tattoos and paintings on state-of-the-art face recognition systems was investigated. To do so, an appropriate database of facial images before and after individuals got facial tattoos or paintings was collected from various online sources. Using five open-source systems and a commercial system, it was shown that facial tattoos and paintings can substantially degrade the performance of these systems. More specifically, it was shown that facial tattoos and paintings generally impact detection algorithms' ability to detect faces and reduce their confidence scores, especially for large facial tattoos and paintings. For face quality estimation using \serfiq and \faceqnet it was shown that the quality score decreases as the size of tattoos and paintings on a face increases. For comparison and feature extraction, it was shown that large facial tattoos and paintings had a big impact on the commercial and especially the open-source systems' ability to extract and compare facial features whereas medium manipulations had some impact. Conclusively, the results show that facial tattoos and paintings negatively affect the biometric performance of face recognition systems, thus indicating a need for developing algorithms robust to such appearance changes.

%% file: sections/acknowledgements.tex
\section*{Acknowledgements}
This research work has been funded by the German Federal Ministry of Education and Research and the Hessian Ministry of Higher Education, Research, Science and the Arts within their joint support of the National Research Center for Applied Cybersecurity ATHENE and the European Union’s Horizon 2020 research and innovation programme under the Marie Skłodowska-Curie grant agreement No. 860813 - TReSPAsS-ETN.